%% file: main.tex
\documentclass[twoside]{article}

\usepackage[accepted]{aistats2022}
\usepackage[round]{natbib}

\input{mathdef.tex}
\input{math_commands.tex}

\usepackage[utf8]{inputenc} 
\usepackage[T1]{fontenc}    
\usepackage{hyperref}       
\usepackage{url}            
\usepackage{booktabs}       
\usepackage{amsfonts}       
\usepackage{nicefrac}       
\usepackage{microtype}      
\usepackage[pdftex,dvipsnames]{xcolor}  
\usepackage{wrapfig,lipsum}
\usepackage{pifont}
\definecolor{myred}{RGB}{215,48,39}
\definecolor{mygreen}{RGB}{26,152,80}
\newcommand{\cmark}{\textcolor{mygreen}{\ding{51}}}
\newcommand{\xmark}{\textcolor{myred}{\ding{55}}}
\usepackage{xspace}
\usepackage{enumitem}

\newcommand*{\ie}{{\it i.e.}\@\xspace}

\renewcommand{\ra}[1]{\renewcommand{\arraystretch}{#1}}
\newcommand{\D}{\mathcal{D}}

\usepackage{mathtools}

\usepackage{floatrow}
\usepackage{microtype}
\usepackage{graphicx}
\usepackage{booktabs} 

\usepackage{wrapfig}
\usepackage{caption}
\usepackage{subcaption}
\usepackage{float}

\usepackage{xargs}                      
\usepackage[colorinlistoftodos]{todonotes}
\newcommandx{\improvement}[2][1=]{\todo[linecolor=Plum,backgroundcolor=Plum!25,bordercolor=Plum,#1]{#2}}

\definecolor{mydarkblue}{rgb}{0,0.08,0.45}
\hypersetup{
    colorlinks=true,
    linkcolor=mydarkblue,
    citecolor=mydarkblue,
    filecolor=mydarkblue,
    urlcolor=mydarkblue
}

\newif\ifcomments

\commentstrue

\ifcomments
\newcommand{\ricky}[1]{\textcolor{blue}{[RC: #1]}}
\else
\newcommand{\ricky}[1]{}
\fi

\ifcomments
\newcommand{\winnie}[1]{\textcolor{blue}{[WX: #1]}}
\else
\newcommand{\winnie}[1]{}
\fi

\ifcomments
\newcommand{\XL}[1]{\textcolor{cyan}{(XL: #1)}}
\else
\newcommand{\XL}[1]{}
\fi

\begin{document}

%
\runningtitle{Infinitely Deep Bayesian Neural Networks with SDEs}

%

\twocolumn[

\aistatstitle{Infinitely Deep Bayesian Neural Networks \\ with Stochastic Differential Equations}

\aistatsauthor{ Winnie Xu \And Ricky T.Q. Chen \And  Xuechen Li \And David Duvenaud }

\aistatsaddress{ University of Toronto \And  University of Toronto \And Stanford Unviersity \And University of Toronto } ]

\begin{abstract}
We perform scalable approximate inference in continuous-depth Bayesian neural networks.
In this model class, uncertainty about separate weights in each layer gives hidden units that follow a stochastic differential equation.
We demonstrate gradient-based stochastic variational inference in this infinite-parameter setting, producing arbitrarily-flexible approximate posteriors.
We also derive a novel gradient estimator that approaches zero variance as the approximate posterior over weights approaches the true posterior.
This approach brings continuous-depth Bayesian neural nets to a competitive comparison against discrete-depth alternatives, while inheriting the memory-efficient training and tunable precision of Neural ODEs.
\end{abstract}

\section{INTRODUCTION}\label{sec:intro}
Taking the limit of neural networks to be the composition of infinitely many residual layers provides a way to implicitly define its output as the solution to an ODE~\citep{haber2017stable,E2017}.
This continuous-depth parameterization decouples the specification of the model from its computation.
While the paradigm adds complexity, it has several benefits:
(1) Computational cost can be traded for precision in a fine-grained manner by specifying error tolerances for adaptive computation, and
(2) memory costs for training can be significantly reduced by running the dynamics backwards in time to reconstruct activations of intermediate states needed for backpropagation.

On the other hand, the Bayesian treatment for neural networks modifies the typical training pipeline where instead of performing point estimates, a distribution over parameters is inferred.
Although this approach adds complexity, it automatically accounts for model uncertainty. In turn, model averaging can be done to combat overfitting and improve calibration, especially on out-of-distribution data~\citep{zhang2018noisy, osawa2019practical}. 


\begin{figure}[t]
    \centering
    \phantom{bl} ODE-Net $\qquad \ \ \ \ \ \ \quad\quad$ SDE-BNN\\
    \begin{subfigure}[b]{\textwidth}
    \includegraphics[trim={0.3cm 0.5cm 0.2cm 0.5cm},clip,width=\linewidth]{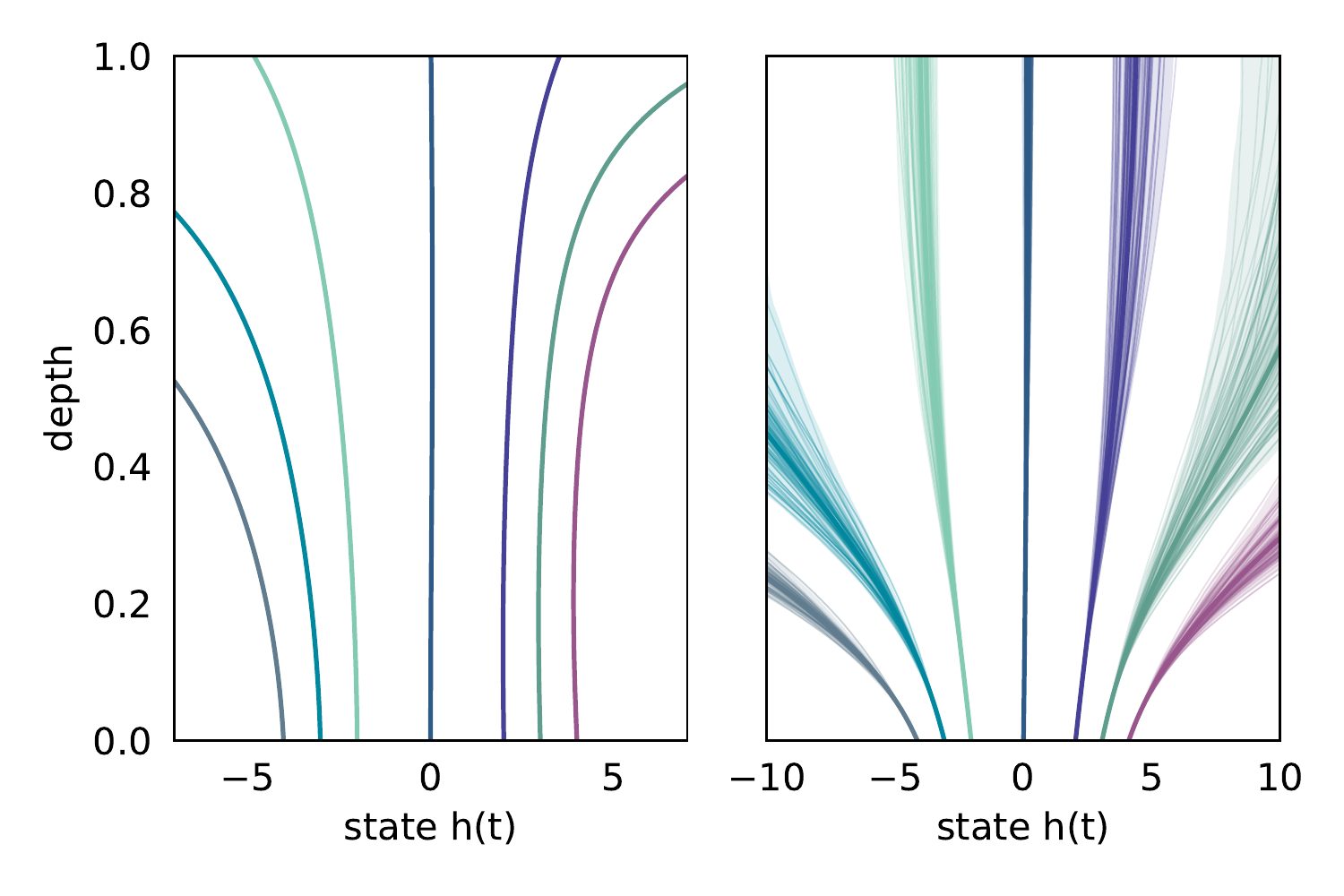}
    \end{subfigure}
\caption{Hidden unit trajectories in an ODE-Net and an SDE-BNN. \textit{Left}: A continuous-depth residual network has deterministic transformations of its hidden units from depths $t=0$ to $t=1$. \textit{Right}: Uncertainty in the weights of a Bayesian continuous-depth residual network implies uncertainty in its hidden unit activation trajectories. Shaded regions show densities over samples from the learned posterior dynamics. \textit{Both}: Each distinct color corresponds to a different initial state corresponding to different data inputs.}

\label{fig:dynamics}
\end{figure}


How can we combine the benefits of continuous-depth models with those of Bayesian neural networks?
The simplest approach is a ``Bayesian neural ODE''~\citep{yildiz2019ode,dandekar2020bayesian}, which integrates out the finitely-many parameters of a standard neural ODE for prediction.

\begin{figure}[t]
    \centering
    \begin{subfigure}[b]{\linewidth}
         \centering
         $\qquad \qquad$ Prior $\ \ \qquad\qquad$ Approximate posterior\\
         \includegraphics[clip, trim=1em 1em 1em 1em, width=\linewidth]{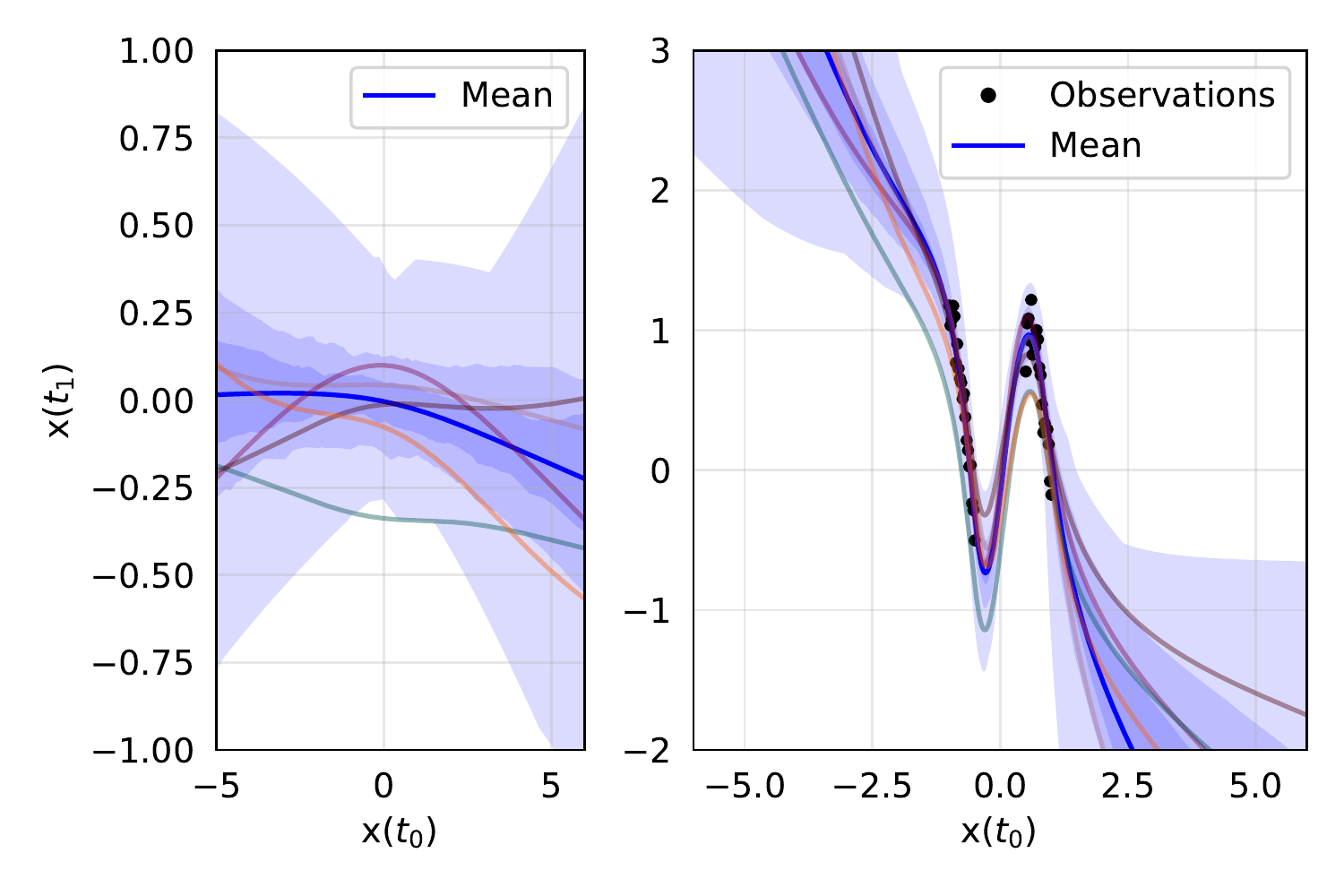}
         \label{fig:toy_prior_posterior}
     \end{subfigure}\\
    \caption{Predictive prior and posterior of the SDE-BNN on a non-monotonic toy dataset.
    Blue areas indicate density percentiles, and distinct colored lines show model samples.
    }
    \label{fig:toypriorposterior}
\end{figure}

This approach is straightforward to implement, and can inherit the advantages of both Bayesian and continuous-depth neural nets.
However, empirically, standard Gaussian approximate posteriors are a relatively poor match for neural ODEs, not to mention the drawbacks of also being used in the prior.
Additionally, it does not exploit the special synergy available between continuous-time models and approximate inference.

In this paper, we show that an alternative construction of Bayesian continuous-depth neural networks has additional practical benefits.
Specifically, we consider the limit of infinite-depth Bayesian neural networks with separate unknown weights at each layer, a model class that we refer to as SDE-BNNs. 
We develop a unique network architecture that enhances model expressivity through time-correlated weights and scales linearly, instead of quadratically, with the parameter dimensionality. Combined with our novel formulation of a zero-variance gradient estimator, we show that approximate inference can be realized through the maximization of our modified variational lower bound, effectively scaling up the gradient-based variational inference scheme described by~\citet{li2020scalable} (preliminary forms of which appeared in earlier works~\citep{archambeau2008variational,opper2019variational,tzen2019neural}).

With this approach,  the state of the output layer is computed by a black-box adaptive SDE solver. 
Figure~\ref{fig:dynamics} contrasts our neural SDE with the 
neural ODE parameterization.
This approach maintains the adaptive computation and constant-memory cost of training Bayesian neural ODEs and adds two unique benefits:
\begin{itemize}[topsep=0pt,leftmargin=*]
    \setlength\itemsep{0.2em}
    \item The variational posterior can be made arbitrarily expressive by simply enlarging the neural network that parameterizes the dynamics of the approximate posterior. Under mild conditions, this approach can approximate the true posterior arbitrarily closely.
    \item The variational objective admits a variance-reduced gradient estimator that is a natural extension of the ``sticking the landing'' trick~\citep{roeder2017sticking}. Combined with arbitrarily expressive approximate posteriors, it is consistent and has vanishing variance as the approximate posterior approaches the true.
\end{itemize}

Notably, our low-variance gradient estimator can also be applied to variational inference in SDEs more generally, such as for time-series modeling, but such applications are beyond the scope of this paper.
\section{BACKGROUND} \label{sec:background}
\paragraph{Bayesian Neural Networks}
Given a dataset, there are often many functions that fit the data well, which a given neural network can express with different parameter values.
Instead of making point estimates of the parameters, the Bayesian paradigm frames learning as posterior inference. Predictions are obtained through integrating over many possible parameter settings. 
Formally, given a dataset $\D= \{(x_i, y_i) \}_{i=1}^N$ and prior distribution over model weights $p(w)$, we want to compute a posterior $p(w | \D) \propto p(\D | w) p(w)$. We can optimize an approximate posterior distribution $q(w)$ that minimizes the Kullback-Leibler (KL) divergence, i.e. maximizing the Evidence Lower Bound (ELBO):
\eq{
\label{eq:elbo}
\! \mathcal{L}_{\text{ELBO}}(\phi) =& 
\E_{q(w)} \sbracks{ \log p(\D| w) } - \KL\bracks{ q(w) || p(w) }.
}
Estimating gradients of this objective using simple Monte Carlo is known as stochastic variational inference (SVI)~\citep{hoffman2013stochastic, rezende2014stochastic}).

One of the main technical challenges of SVI is choosing a parametric family of approximate posteriors that is tractable to sample from and evaluate, while being flexible enough to approximate the true posterior well. 
Most scalable inference techniques use Gaussian approximate posteriors with restricted covariance structure between network parameters
~\citep{graves2011practical,blundell2015weight,zhang2018noisy,mishkin2018slang}. 
Others construct complex approximate posteriors with normalizing flows~\citep{krueger2018bayesian,louizos2017multiplicative} or through distillation~\citep{balan2015bayesian,wang2018adversarial}.

\paragraph{Neural Ordinary Differential Equations}

\begin{figure}
    \centering
    \begin{subfigure}[b]{0.05\linewidth}
        \rotatebox{90}{$z(t)$} \vspace{5em}
    \end{subfigure}
    \begin{subfigure}[b]{0.37\linewidth}
    \includegraphics[trim={1.8cm 0.5cm 0 0},clip,width=\linewidth]{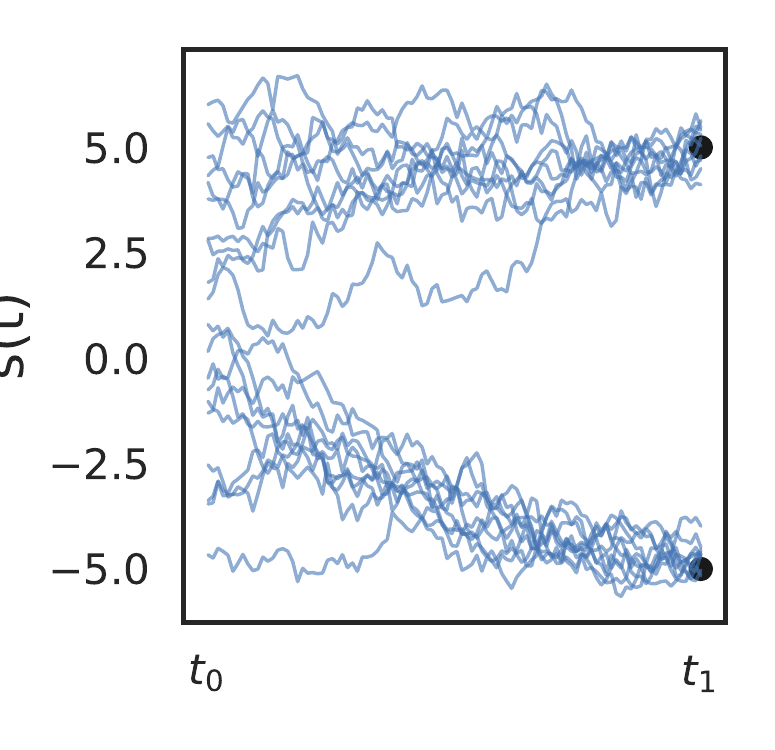}
    \hspace{-2em}
    \vspace{0.5em}
    \phantom{blablab} $t$
    \end{subfigure}
    \begin{subfigure}[b]{0.05\linewidth}
    \rotatebox{90}{$z(t_1)$} \vspace{4.8em}
    \end{subfigure}
    \begin{subfigure}[b]{0.46\linewidth}
    \includegraphics[trim={1.5cm 1.5cm 0 0},clip,width=0.95\linewidth]
    {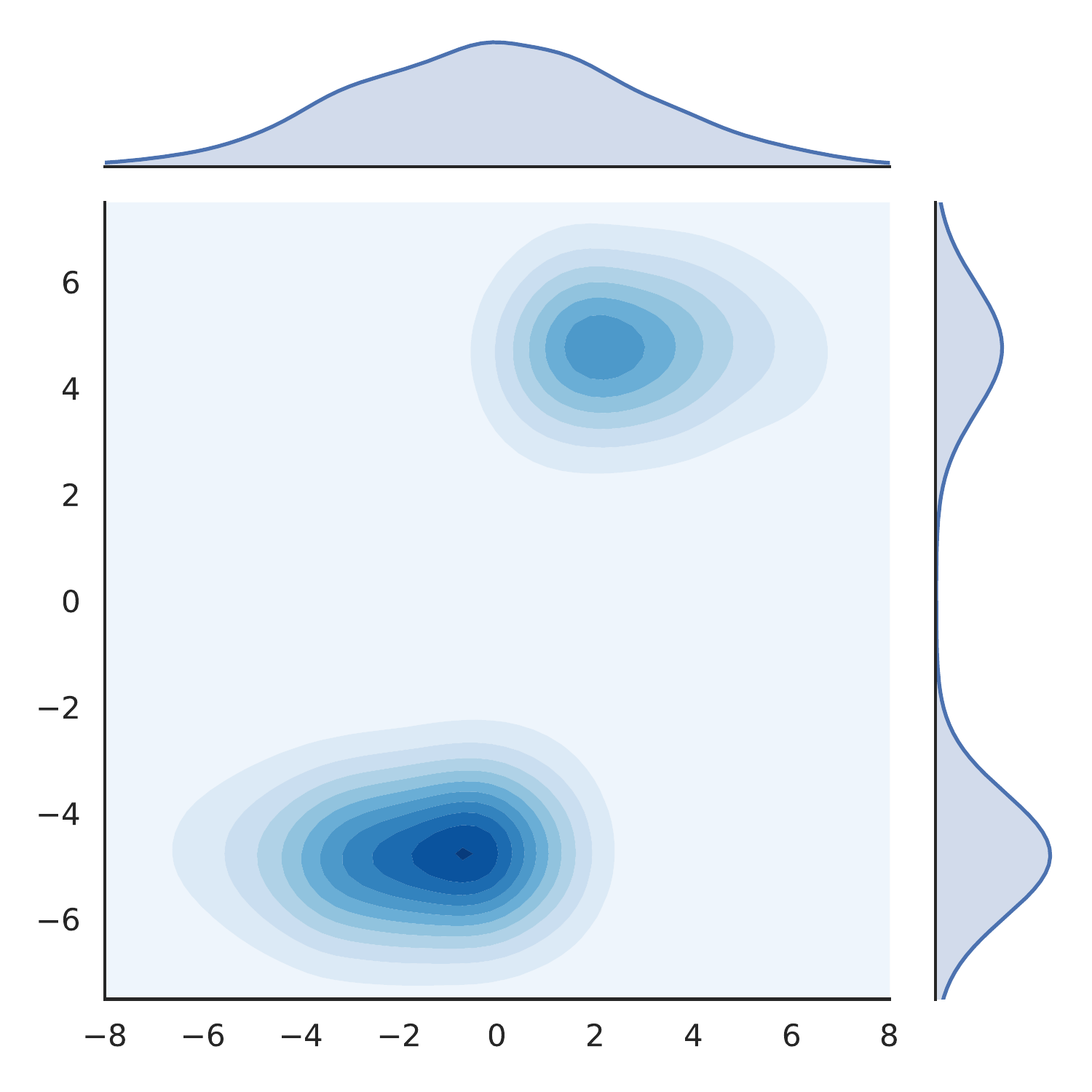}\\
    \hspace{-2em}
    \vspace{0.5em}
    \phantom{blabl} $z(t_0)$
    \end{subfigure}
\caption{Neural SDEs can learn arbitrarily expressive approximate posteriors.
\emph{Left:} Samples from an approximate posterior, trained with an OU prior and conditioned on two observations with Cauchy likelihoods.
\emph{Right:} Joint distribution and marginals of the approximate posterior process $z$ at times $t_0$ and $t_1$.}
\label{fig:multimodalcauchy}
\end{figure}


Neural ordinary differential equations~\citep{chen2018neural} define ODEs using neural networks:
\eq{
\dee h_t = f_\theta(h_t, t) \dt, \quad h_0 \in \R^d,
}
where $f:\R^d \times \R \to \R^d$ is a Lipschitz function defined by a neural network with parameters $\theta$.
%
Starting at an initial value $h_0 = x$ given by a data example and integrating these dynamics forward for a finite time can be seen as passing the input through an infinitely-deep residual network.
For learning scalar-valued functions, adding extra dimensions to $h$ and a linear final layer induces similar universal approximations to 
standard neural networks~\citep{dupont2019augmented,zhang2019approximation} trained by standard stochastic gradient descent methods.
Using adaptive ODE solvers can trade evaluation speed for precision. The adjoint sensitivity method saves memory during training through reconstructing the trajectory of the hidden units $h$ by running the dynamics backwards in backpropagation.
\subsection{Latent Stochastic Differential Equations}\label{ref:latentsde}
Informally, an SDE can be viewed as an ODE with infinitesimal noise added throughout time. Formally:
\begin{align}
\dee w_t = f_\theta(w_t, t)\dt+ g_\theta(w_t, t) \dB_t,
\label{eq:prior}
\end{align}
where $w_0 \in \R^d$ is the initial state, 
$f_\theta:\R^d \times \R \to \R^d$ and $g_\theta:\R^d \times \R \to \R^{d \times m}$ are functions Lipschitz in both arguments, dubbed the \textit{drift} and \textit{diffusion}, respectively, and $\{B_t\}$ is an $m$-dimensional Brownian motion.

Some works have considered training SDEs with dynamics parameterized by neural networks~\citep{li2020scalable,tzen2019neural,peluchetti2020infinitely,innes2019zygote,kong2020sde,liu2019neural}. 
Note that directly optimizing the drift and diffusion to maximize the average log-likelihood of an observation $\log p(y_t | w_t)$ would result in the diffusion approaching $0$, conditional on the ODE fitting the training data well with the diffusion somewhat unconstrained.

Instead of directly optimizing the parameters of an SDE to match the data, a better approach is to use an SDE to define a prior over trajectories of $w$, and optimize the marginal likelihood of the data, integrating over all trajectories of $w$ weighted by the prior.
%
Luckily, we can specify an approximate posterior over trajectories using a second SDE.
We define the approximate posterior by
\begin{align}
\dw_t &= f_\phi(w_t, t) \dt + g_\theta(w_t, t) \dB_t. \label{eq:posterior}
\end{align}
When the dynamics of the approximate posterior $f_\phi$ is parameterized by a neural network, this family of  posteriors is extremely expressive.
Figure~\ref{fig:multimodalcauchy} shows that such a variational family can easily approximate non-Gaussian and multi-modal posteriors on path space.

If both the SDE defined by~\eqref{eq:prior} and~\eqref{eq:posterior} share the same diffusion function, then the KL between the two induced measures on path space has the following form~\citep{li2020scalable,tzen2019neural}:
\vspace{-1.7em}
\begin{align}
\KL \bracks{ \mu_q || \mu_p } = 
    \mathbb{E}_{q_\phi(w)}\left[
        \int_0^1 \!\!\! \tfrac{1}{2} \norm{ u(t, \phi) }_2^2 \dt \right] \label{eq:kl} \textnormal{where} \\ 
     \qquad u(t, \phi) = g_\theta(w_t, t)^{-1} \left[f_\theta(w_t, t) - f_\phi(w_t, t) \right] \label{eq:u}
\end{align}

where $\mu_q$ and $\mu_p$ are path space probability measures induced respectively by~\eqref{eq:posterior} and~\eqref{eq:prior}, and the expectation is taken under the approximate posterior, denoted $q_\phi(w)$.
Intuitively, this KL divergence resembles the summative difference over time horizon [0, 1] between the prior drift $f_\theta$ and $f_\phi$, scaled by the diffusion.
This divergence can be estimated up to a constant with simple Monte Carlo, sampling trajectories from the dynamics given by the approximate posterior.
\paragraph{SDEs as expressive approximate posteriors}
To ensure that the KL divergence between the prior and approximate posterior on path space is finite, the same diffusion function $g_\theta(w_t, t)$ must be used for the approximate posterior and prior.
Surprisingly, this does not limit the expressivity of the approximate posterior. 
\citet{boue1998variational} show that there is a one-to-one correspondence between the space of path measures and drift functions that result in the same path space KL divergence. 
This implies that any path space measure close to the true posterior can be instantiated by SDEs with appropriate drifts. It follows that an approximate posterior parameterized by a sufficiently expressive family of function approximators can be made arbitrarily close to the true posterior.
The Girsanov reparameterization of the variational formula, derived from Boue \citep[Section 4]{tzen2019neural}, proves that the ELBO is tight when the drift is optimal. This means that there exists a ground truth drift function that can make the ELBO tight, the approximation of which can be achieved with a high capacity neural network.

\section{INFINITELY DEEP BNNs}\label{sec:infinite bnns}
Standard discrete-depth residual networks can be defined as a composition of layers of the form:
\begin{align}
    h_{t+\epsilon} = h_t + \epsilon f(h_t, w_t),  \quad t = 1 \dots T,
\end{align}
where $t$ is the layer index, $h_t \in \mathbb{R}^{D_h}$ denotes a vector of hidden unit activations at layer $t$, the input $h_0 = x$, and $w_t \in \mathbb{R}^{D_w}$ represents the parameters for layer $t$.
In the discrete setting, $\epsilon = 1$, $\in\mathbb{R}$.

We can construct a continuous-depth variant of residual networks by setting $\epsilon = \nicefrac{1}{T}$ and taking the limit as $T \rightarrow \infty$.
This yields a differential equation that describes the hidden unit evolution as a function of depth $t$.
Since standard residual networks are parameterized with different layerwise ``weights'', we denote them $w_t$.
To specify different weights at each layer with a finite number of parameters, we introduce a hypernetwork $f_w$ that specifies the change in weights as a function of depth and the current weights~\citep{ha2016hypernetworks}.
The evolution of the hidden unit activations and weights can then be combined into a single differential equation: 
\begin{align}
\frac{\dee}{\dt}
\begin{bmatrix}
h_t \\
w_t
\end{bmatrix} = 
\begin{bmatrix}
f_h(t, h_t, w_t) \\
f_w(t, w_t)
\end{bmatrix}
\end{align}
with some learned initial weight value $w_{t_0}$. Using time-varying weights is similar to augmenting the state~\citep{dupont2019augmented,anodev2}. See Appendix Figure~\ref{fig:posteriorpredictive} on the effects of augmentation.
We perform Bayesian inference on the weight process $w_t$, assigning a suitable prior stochastic process and performing variational inference in this infinitesimal limit.

Like all Bayesian neural networks with observation likelihoods, our framework models uncertainty both about parameters and about individual observations:
The likelihood $p(y | h_1)$ captures the observational noise, while the SDE encodes weight uncertainty. 
\paragraph{Prior process on weights}
Typical priors for Bayesian neural networks are independent Gaussians across all weights and layers.
Taking the infinitesimal limit of such a prior gives a white noise process prior on the weights $w(\cdot)$.
However, initializing this noise while maintaining finite variance at scale is difficult~\citep{peluchetti2020doubly,peluchetti2020infinitely}.

Instead, we use the Ornstein–Uhlenbeck (OU) process as the prior on weights.
The process is characterized by an SDE with drift and diffusion:
\begin{align}
    f_p(w_t, t) = - w_t, \quad g(w_t, t) = \sigma I_d,
\end{align}
respectively, where $\sigma$ is a hyperparameter.
We choose this prior for its simplicity and bounded marginal variance at a constant in the large time limit.
\paragraph{Approximate posterior over weights}
We parameterize the approximate posterior on weights implicitly using another SDE with the following drift function:
\begin{align}
    f_q(w_t, t, \phi) &= \text{NN}_\phi(w_t, t, \phi) - f_p(w_t, t).
\end{align}
This drift $f_q$ is parameterized by a small neural network (NN) with parameters $\phi$.
With this drift, the approximate posterior process will generally have non-Gaussian, non-factorized marginals; its expressive capacity can be increased by making the neural net larger.

\paragraph{Evaluating the network}
Given an input, we marginalize over weight and hidden unit trajectories.
This can be done with simple Monte Carlo, sampling a weight path $\{w_t\}$ from the posterior process and evaluating the network activations $\{h_t\}$ given the sampled weights and input.
Both steps require solving a differential equation.
Luckily, both can be solved simultaneously with the augmented state SDE:
\begin{align}\label{augsde}
    \dee \begin{bmatrix}
    w_t \\
    h_t
    \end{bmatrix} = 
    \begin{bmatrix}
    f_w(w_t, t, \phi) \\
    f_h(h_t, t,  w_t)
    \end{bmatrix}  \dt + 
    \begin{bmatrix}
    g_w(w_t, t)\\
    \mathbf{0}
    \end{bmatrix} \dB_t, 
\end{align}
where $h_0 = x$, the input. 
The learnable parameters are the initial weight values at time zero $w_0$ (either point estimated or inferred) and those of drift function $\phi$.
\paragraph{Output likelihood}
The final state of the hidden units $h_1$ is used to parameterize the likelihood of the target output $y$:
 $\log p(y | x, w) = \log p(y | h_1)$.
For instance, $p(y | h_1)$ could be a Cauchy likelihood for regression, or categorical likelihood for classification.
\paragraph{Training objective}
To fit the network to data, we maximize the lower bound on marginal likelihood given by the infinite-dimensional ELBO:
\begin{equation}
\mathcal{L}_{\text{ELBO}_\infty}(\phi)  
= \mathbb{E}_{q_\phi(w)} \left[
    \log p(\mathcal{D} | w) \!-\!\!
    \int_0^1 \!\! \tfrac{1}{2}
        \norm{u(w_t, t, \phi)}_2^2 \dt \right]. \nonumber
    \nonumber \label{eq:sde_elbo}
\end{equation}
The sampled weights, the hidden activations, and the training objective are all computed simultaneously with a single call to an adaptive SDE solver.
Gradients of the sampled loss can also be efficiently computed using adaptive solvers, following \citet{li2020scalable}.

\section{VARIANCE-REDUCED GRADIENTS}\label{sec:stl}
\begin{figure}[t]
    \centering
    \begin{subfigure}[b]{\linewidth}
    \centering
    \includegraphics[width=\linewidth, trim=2px 2px 0 2px, clip]{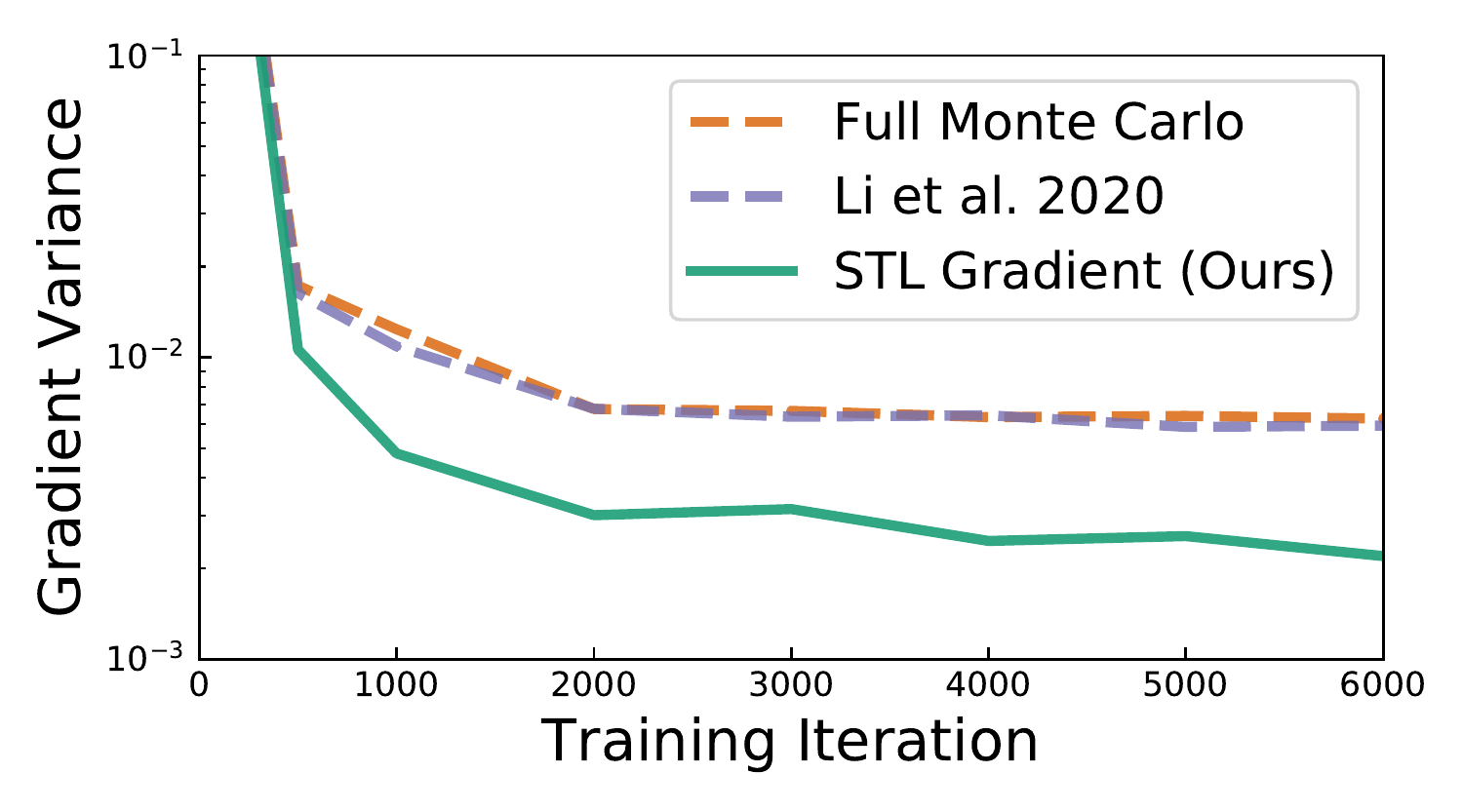}
    \end{subfigure}
    \caption{Comparison of the variance in three gradient estimators. On this toy problem, our new gradient estimator reduces variance by a factor of roughly 4.}
    \label{fig:toy_stl}
\end{figure}

\citet{roeder2017sticking} showed that when optimizing expectations using the reparameterization gradient, a gradient estimator with lower variance can be constructed by removing a score function term that has zero expectation, and that the variance of this gradient estimator approaches zero as the approximate posterior approaches the true posterior.
We refer to this trick as ``sticking the landing'' (STL). 
We generalize this to our SDE setting by replacing the original estimator of the path space KL with the following STL estimator:
\eq{
\widehat{\text{KL}}_{\text{STL}} \!=\! 
    \int_0^1
        \!\! \tfrac{1}{2} \norm{ u(w_t, t, \phi) }_2^2 \dt \! + \!\!
    \int_0^1
        \!\! u(w_t, t, \bot(\phi)) \dB_t, 
\label{eq:stl}
}
where $w(\cdot) \sim q_\phi(w)$ and $u$ is defined in \eqref{eq:u}, the path $\{w_t\}_{t \in [0, T]}$ is sampled from the approximate posterior process, and $\bot(\cdot)$ is the stop gradient function that renders the input a constant with respect to which gradient propagation is stopped. Note that \ref{eq:stl} is the fully Monte Carlo version referred to in \ref{fig:toy_stl} from which our STL variant is derived.

The second term in~\eqref{eq:stl} is a martingale and has expectation zero. Therefore, in prior works~\citep{li2020scalable,tzen2019neural,tzen2019theoretical}, Monte Carlo estimation was only performed for the first term, but we find that this approach does not necessarily reduce the variance of the \emph{gradient}~(Figure~\ref{fig:toy_stl}).

Because our approximate posterior can be made arbitrarily expressive, we conjecture that our approach can achieve arbitrarily low gradient variance towards the end of training if the $f_w$ parameterization is expressive enough. See Appendix~is \ref{app:stlheuristic} for a heuristic derivation.

We show the variance of different gradient estimators in Figure~\ref{fig:toy_stl}, averaged across the parameters $\theta$, in a 1D regression setting. We compare STL against a ``Full Monte Carlo'' estimate which includes the second additional term without gradient stopping, as well as the estimator that was previously used by~\citet{li2020scalable} which ignores the second term.
Figure~\ref{fig:toy_stl} shows that STL obtains lower variance than alternatives, when matching an exponentiated Brownian motion.
Table~\ref{tab4:stl_improvement} shows training performance improvements.
\section{EXPERIMENTS}\label{sec:experiments}
We investigate the effectiveness of our proposed approximate inference method for training continuous-depth neural nets, referred to as SDE-BNN, in terms of classification accuracy, calibration, perturbation robustness, and speed-precision trade-offs. 
Our code is publicly available
\href{https://github.com/xwinxu/bayeSDE}{here} and experimental settings in Table~\ref{tab:hparam settings}.

\newcommand{\tcen}[1]{\multicolumn{1}{c}{#1}}
\newcommand{\tnan}{---}
\begin{table*}\centering
\ra{1.1}
\setlength{\tabcolsep}{0.8em}
\resizebox{0.75\textwidth}{!}{
\begin{tabular}{@{} l c c c c c c @{}}\toprule
  & \multicolumn{2}{c}{\textbf{\textsc{MNIST}}} &
  & \multicolumn{2}{c}{\textbf{\textsc{CIFAR-10}}} \\
\cmidrule(lr){2-3} \cmidrule(l){5-6} 
Model & \tcen{Accuracy (\%) } & \tcen{ECE ($\times 10^{-2}$)}
& & \tcen{Accuracy (\%)} & \tcen{ECE ($\times 10^{-2}$)} \\
\cmidrule(r){1-1}\cmidrule(lr){2-3} \cmidrule(l){5-6} 
ResNet32 & 99.46 $\pm$ 0.00 & 2.88 $\pm$ 0.94 & & 87.35 $\pm$ 0.00 & 8.47 $\pm$ 0.39 \\
ODEnet & 98.90 $\pm$ 0.04 & 1.11 $\pm$ 0.10 & & 88.30 $\pm$ 0.29 & 8.71 $\pm$ 0.21 \\
HyperODEnet & 99.04 $\pm$ 0.00 & 1.04 $\pm$ 0.09 & & 87.92 $\pm$ 0.46 & 15.86 $\pm$ 1.25 \\
\cmidrule(r){1-1}\cmidrule(lr){2-3} \cmidrule(l){5-6} 
MFVI ResNet32 & 99.44 $\pm$ 0.00 & 2.76 $\pm$ 1.28 & & 86.97 $\pm$ 0.00 & 3.04 $\pm$ 0.94 \\
MFVI$^\dagger$ & ---  & --- & & 86.48 \phantom{$\pm$ 0.00} & 1.95 \phantom{$\pm$ 0.00} \\
Deep Ensemble$^\dagger$ & --- & --- & & 89.22 \phantom{$\pm$ 0.00} & 2.79 \phantom{$\pm$ 0.00} \\
HMC (``gold standard'')$^\dagger$ & 98.31 \phantom{$\pm$ 0.00}  & 1.79 \phantom{$\pm$ 0.00} & & 90.70 \phantom{$\pm$ 0.00} & 5.94 \phantom{$\pm$ 0.00} \\ 
\cmidrule(r){1-1}\cmidrule(lr){2-3} \cmidrule(l){5-6} 
MFVI ODEnet & 98.81 $\pm$ 0.00 & 2.63 $\pm$ 0.31 & & 81.59 $\pm$ 0.01 & 3.62 $\pm$ 0.40 \\ 
MFVI HyperODEnet & 98.77 $\pm$ 0.01 & 2.82 $\pm$ 1.34 & & 80.62 $\pm$ 0.00 & 4.29 $\pm$ 1.10 \\
SDE BNN & 99.30 $\pm$ 0.09 & 0.63 $\pm$ 0.10 & & 89.84 $\pm$ 0.94 & 7.19 $\pm$ 0.37 \\
SDE BNN (+ STL) & 99.10 $\pm$ 0.09 & 0.78 $\pm$ 0.12 & & 89.10 $\pm$ 0.45 & 7.97 $\pm$ 0.51 \\
\bottomrule
\end{tabular}
}
\caption{Classification accuracy and expected calibration error (ECE) on MNIST and CIFAR-10. 
We separate models into point estimates, discrete-time models, and continuous-time models.
Our SDE-BNN outperforms other continuous-time Bayesian neural nets (BNNs) and perform competitively against discrete-time BNNs.
$^\dagger$Results by \citet{izmailov2021bayesian} where a modified residual network architecture was used; only one seed was reported.
}
\label{tab1:img_results}
\end{table*}

\begin{figure}
    \centering
    \vspace{0em}
    \begin{subfigure}[b]{0.9\linewidth}
         \centering
         \includegraphics[width=\linewidth]{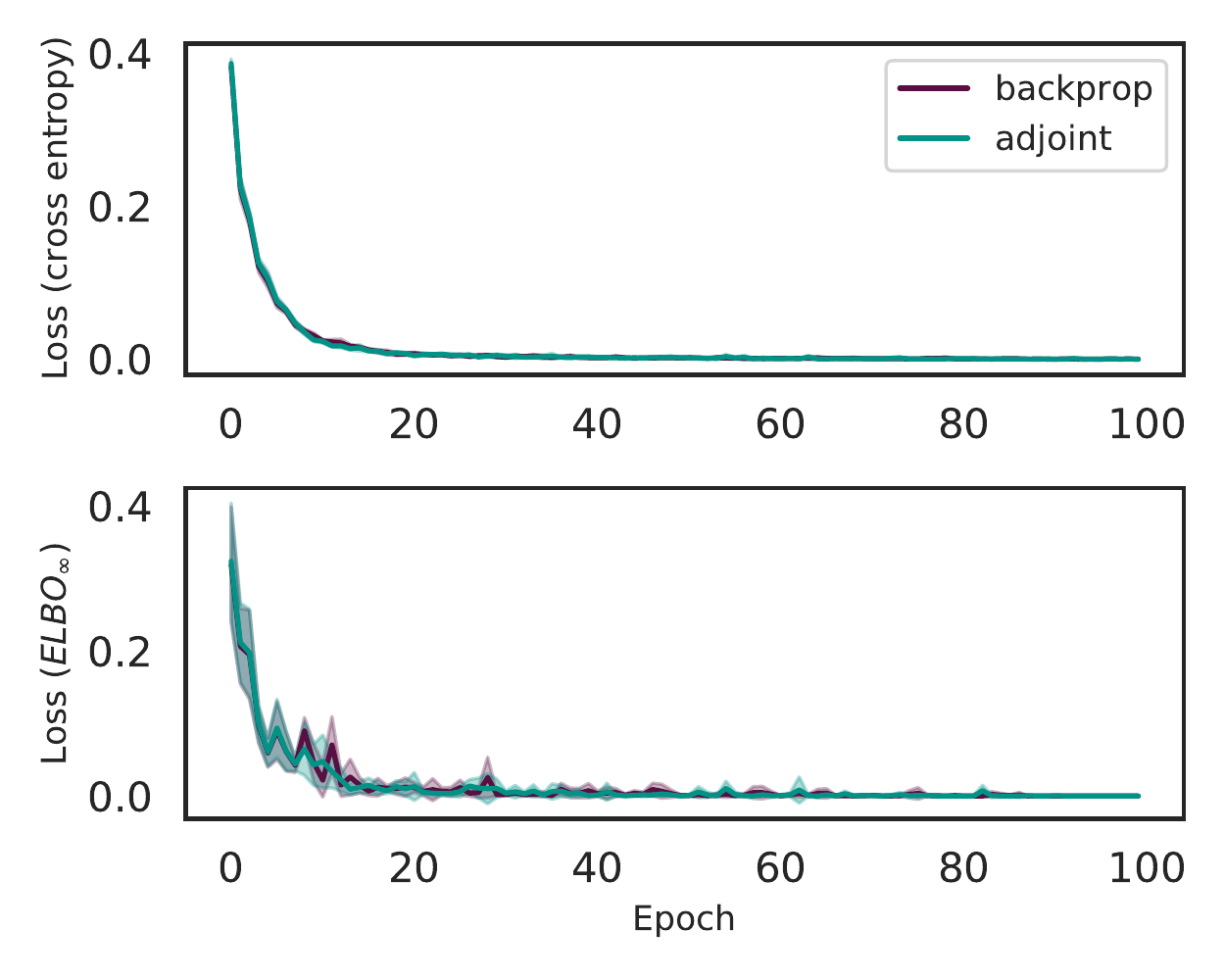}
     \end{subfigure}\\
    \caption{Benchmarking two gradient computation methods: (1) Back-propagation through the SDE solver, and (2) the memory-efficient stochastic adjoint of \citet{li2020scalable}.
    Both methods have similar optimization dynamics, final performance, and wall-clock time, but the adjoint approach is more memory-efficient.
    Detailed comparisons of wall-clock time and evaluation step results in Appendix~\ref{app:adjointadaptivesolver}.}
    \label{fig:bpadjoint}
\end{figure}

We consider toy regression and image classification tasks on MNIST and CIFAR-10.
We also investigate out-of-distribution generalization.
Notably, our approach does not require \emph{post hoc} recalibration methods such as training with temperature scaling \citep{guo2017calibration} or isotonic regression \citep{zadrozny2002transforming}.
\paragraph{Backpropagation through solvers vs. adjoint}
We experimented with fixed- and adaptive-step SDE solvers, and the stochastic adjoint of~\citet{li2020scalable}.
Figure~\ref{fig:bpadjoint} shows similar convergence for both approaches.  Appendix~\ref{app:additional_figures} shows that both had similar numbers of dynamics function evaluations and wall-clock time.


The overhead for estimating error in our adaptive solvers was substantial; therefore, for final model evaluation, we trained with fixed-step solvers, where the number of steps is chosen to be large enough to match the convergence speed of our adaptive-step solvers.

\paragraph{Baselines}
For a fixed-depth network baseline, we compare to standard residual networks.
We then test variational inference on the weights of these models.

We also perform ablation studies to compare with standard variational inference approaches over continuous-depth networks.
Specifically, we compare to a mean field variational inference (MFVI) ODEnet where stochastic variational inference is performed over depth-invariant weights.
This baseline is a fully-factorized Gaussian approximate posterior, \ie mean-field approximation, and been used for Neural ODEs by \citet{look2019differential,dandekar2020bayesian}.

We further compare our model to a MFVI HyperODEnet, where a learned drift is applied to $w$, but mean-field inference is instead performed over the parameters of the hypernetwork. 
Alternatively, one can interpret this as another MFVI ODEnet with a larger state and a more complex drift function but with similar computational complexity to SDE-BNN.
This setting contrasts our approach of doing Bayesian inference over the entire continuous-depth network as a stochastic process. 
\vspace{-1em}
\paragraph{Parameterizing the drift function} 
\vspace{-1em}
We parameterized the drift of the variational posterior $f_w$ with a simple multilayer perceptron.
To ensure optimization starts at a stable set of dynamics, we subtract the prior drift so that the approximate posterior equals the prior when the final layer is initialized to output zero.

\paragraph{Hyperparameters}
We swept learning rates in the range [1e-4, 1e-3], selecting the optimal based on the validation set. We train with the default Adam optimizer~\citep{Kingma2015AdamAM}.
In image classification experiments, all convolutional layers of the drift network are time-conditional and use the \texttt{tanh} non-linearity. The diffusion coefficient $\sigma$ was selected from validation performance over \{0.1, 0.2, 0.5\}.

\subsection{1D Regression}
We first verify the capabilities of the SDE-BNN on a 1D regression problem. 
Conditioned on a sample from the diffusion process, each sample from a one-dimensional SDE-BNN is a bijective mapping from the inputs to the outputs.
This implies that every function sampled from a 1D SDE-BNN is monotonic.
To be able to sample non-monotonic functions, we augment the state with 2 extra dimensions initialized to zero, as in~\citet{dupont2019augmented}.
Figure~\ref{fig:toypriorposterior} shows that our model learns a reasonably flexible approximate posterior on a synthetic non-monotonic 1D dataset.
We emphasize that the samples from our model are smooth w.r.t. depth because the hidden states $h$ do not receive additive instantaneous noise, only the weights $w$ do.
\subsection{Image Classification}
\begin{figure}[htb]
    \centering
    \begin{subfigure}[b]{\textwidth}
        \centering
        \includegraphics[width=0.475\linewidth,trim={1.15cm 0.45cm 0.53cm 0.7cm},clip]{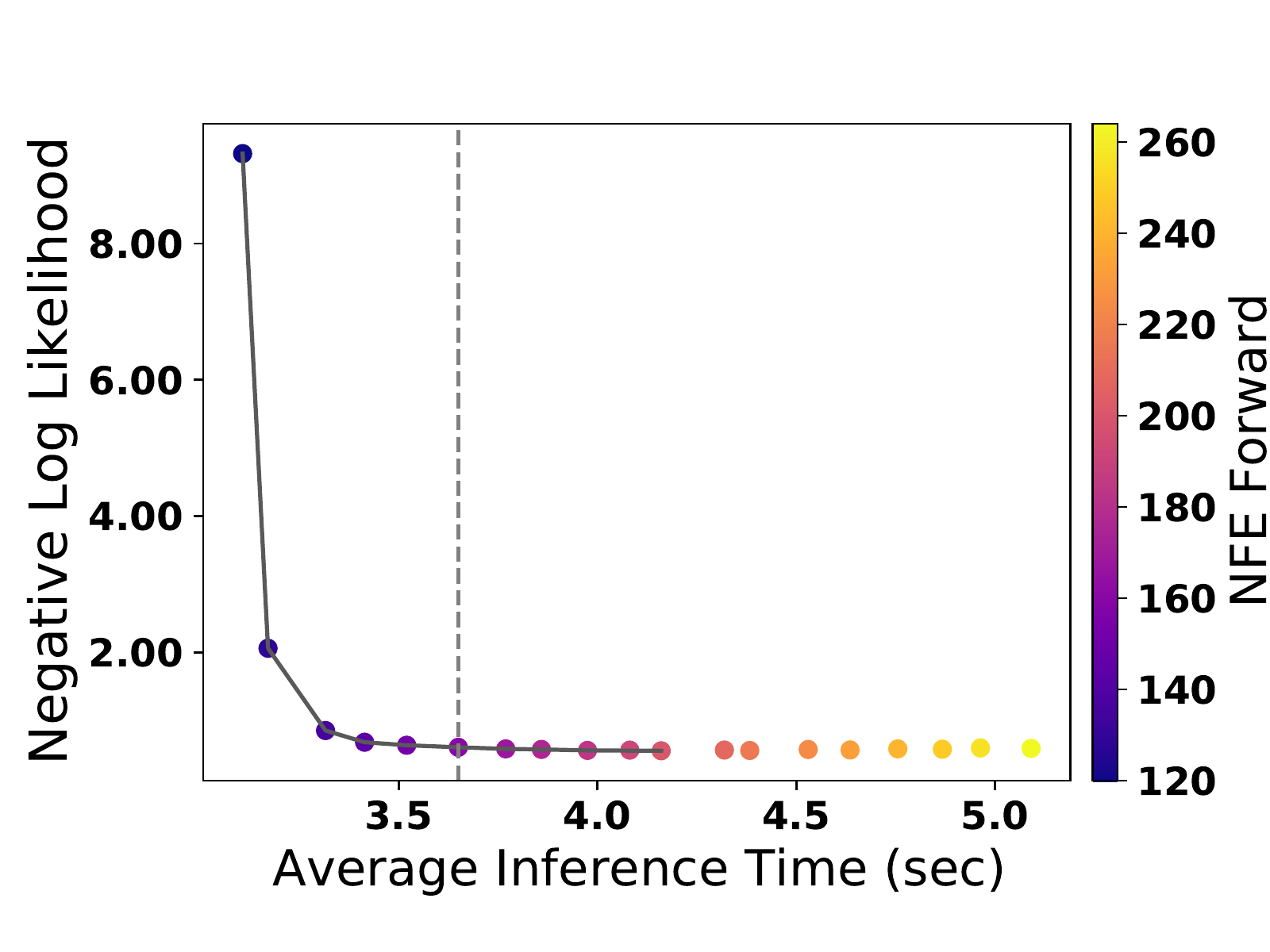}%
        \hfill
        \includegraphics[width=0.475\linewidth,trim={1.17cm 0.42cm 0.3cm 0.7cm},clip]{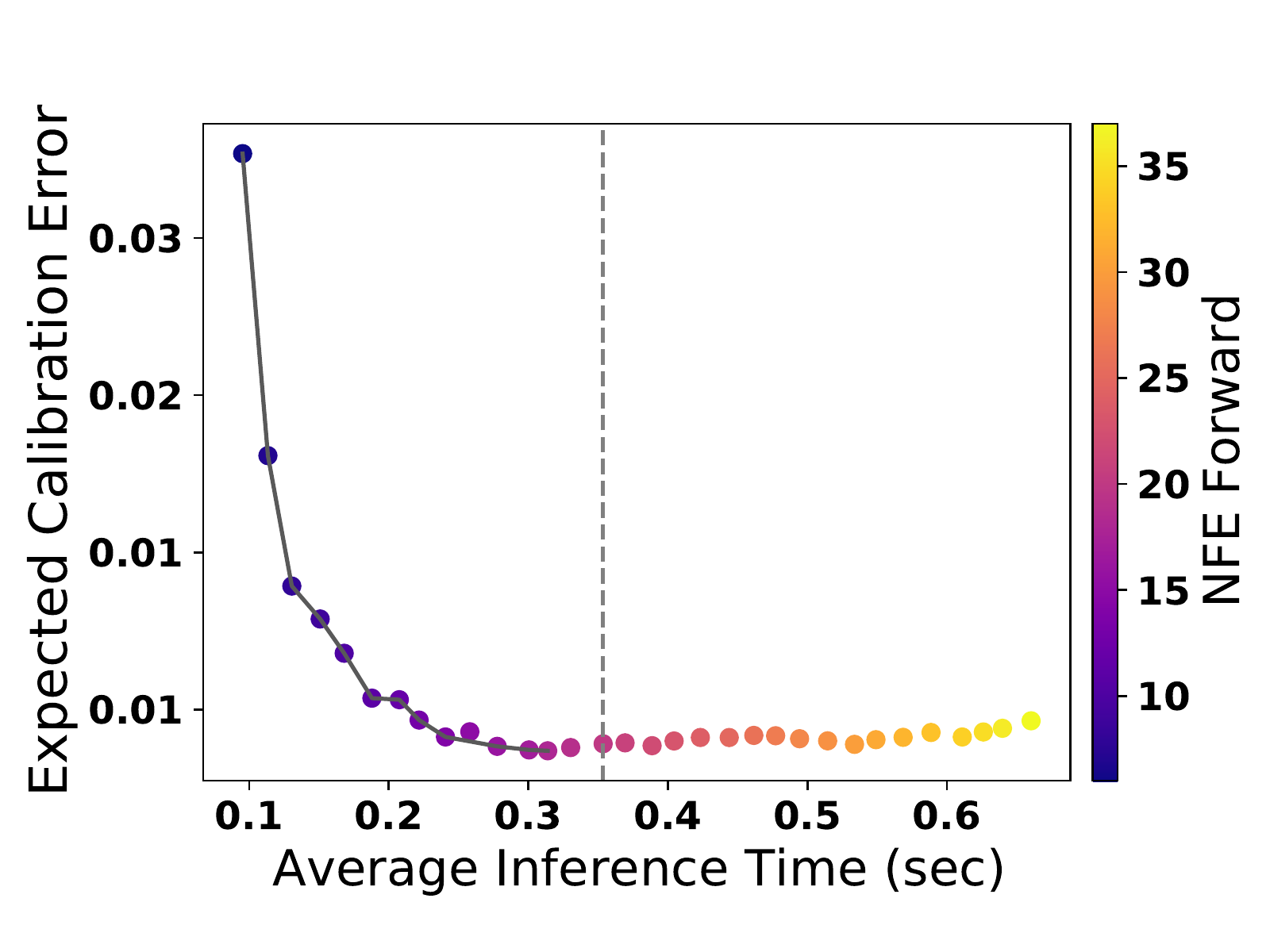}
        \caption{CIFAR-10. \textit{Left:} Negative log likelihood. \textit{Right:} ECE. Adjusting SDE-BNN solver tolerance at test time trades off computational speed for predictive performance. Grey line is solver's training tolerance. Averaged across 3 seeds.}
        \label{fig:pareto}
    \end{subfigure}
    \hspace{0.3em}
    \begin{subfigure}[b]{\textwidth}
        \centering
        \includegraphics[width=0.49\linewidth]{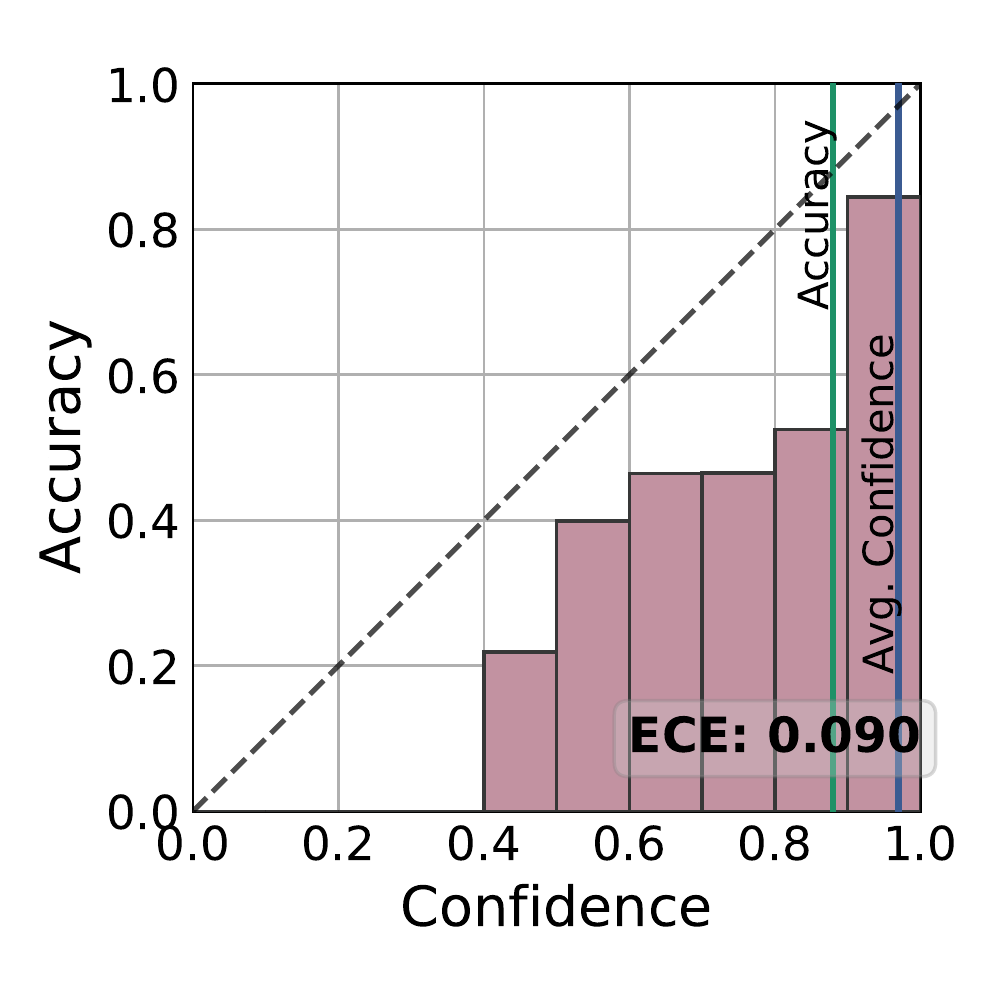}%
        \hspace{0.1em}
        \includegraphics[width=0.49\linewidth]{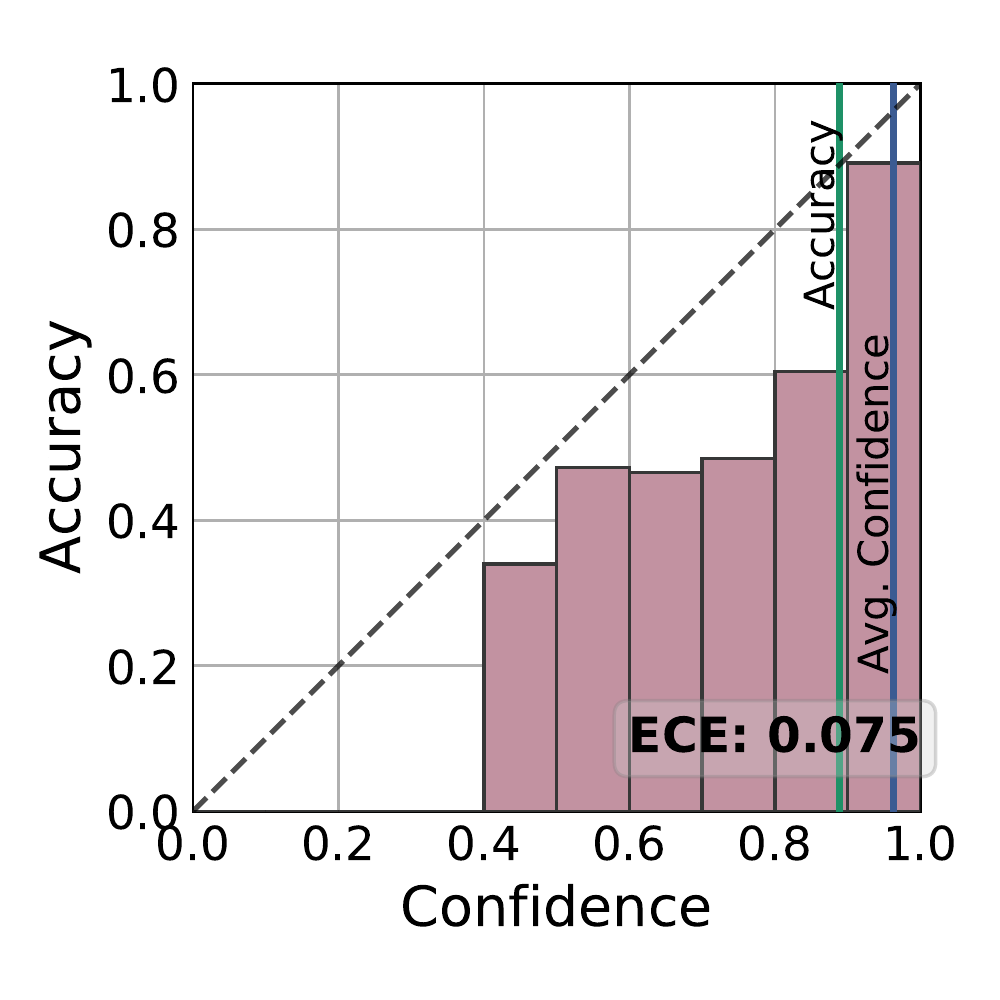}
        \caption{Calibration on the CIFAR-10 test set for a neural ODE (left) and a SDE-BNN (right). The SDE-BNN displays better calibration and generalization.}
        \label{fig:cifarcalibration}
    \end{subfigure}
    \caption{Performance of SDE-BNN on standard CIFAR-10 classification task.}
    \label{fig:imgclassplots}
\end{figure}
Instantaneous changes to the hidden state ($f_h$) are parameterized using a convolutional neural network, including one strided convolution for downsampling and a transposed convolution layer for upsampling. 
We then set the $w$ to be the filters and biases of all the convolutional layers.
The approximate posterior drift dynamics ($f_w$) is a multilayer perceptron with hidden layer widths of 2, 128, and 2. The small hidden width of the bottleneck layers was chosen to reduce the number of variational parameters and promote linear scaling with respect to the dimension of $w$.
On MNIST, we used one such SDE-BNN block, while on CIFAR-10, we used a multi-scale variant where multiple SDE-BNN blocks were stacked with the invertible downsampling from \citet{dinh2016density} in between.

We report classification results in Table~\ref{tab1:img_results}.
Our SDE-BNN generally outperforms the baselines. While the continuous-depth Neural ODE (ODEnet) models can achieve similar classification performance on a standard residual network, it consistently has poorer calibration. 

The SDE-BNN matches and outperforms the accuracy of standard residual networks on MNIST and CIFAR-10, respectively, while obtaining lower expected calibration errors (ECE).
From ablation studies, we found that it was harder to achieve similar performance with either of the mean field variants of an ODEnet as they had a poorer trade-off between performance and calibration.

Figure~\ref{fig:pareto} shows the ability of SDE-BNNs to trade off computation time for precision. 
Figure~\ref{fig:cifar10_calib_spectrum_inference} in Appendix~\ref{app:calibration}  indicates that calibration is insensitive to solver tolerances close to the value used during training.

\subsubsection{Calibration}
Table~\ref{tab1:img_results} quantifies our model's calibration with expected calibration error (ECE;~\citet{guo2017calibration}). 
The SDE-BNN appears better calibrated than the Neural ODE \citep{chen2018neural} and mean field ResNet baselines.
Figure~\ref{fig:cifarcalibration} shows better calibration than neural ODEs with similar accuracy.
Appendix Figure~\ref{fig:cifar10_calib_spectrum} shows the insensitivity of these results to solver step size.
\begin{figure}
    \centering
    \begin{subfigure}[b]{0.8\linewidth}
         \centering
         \includegraphics[width=\linewidth]{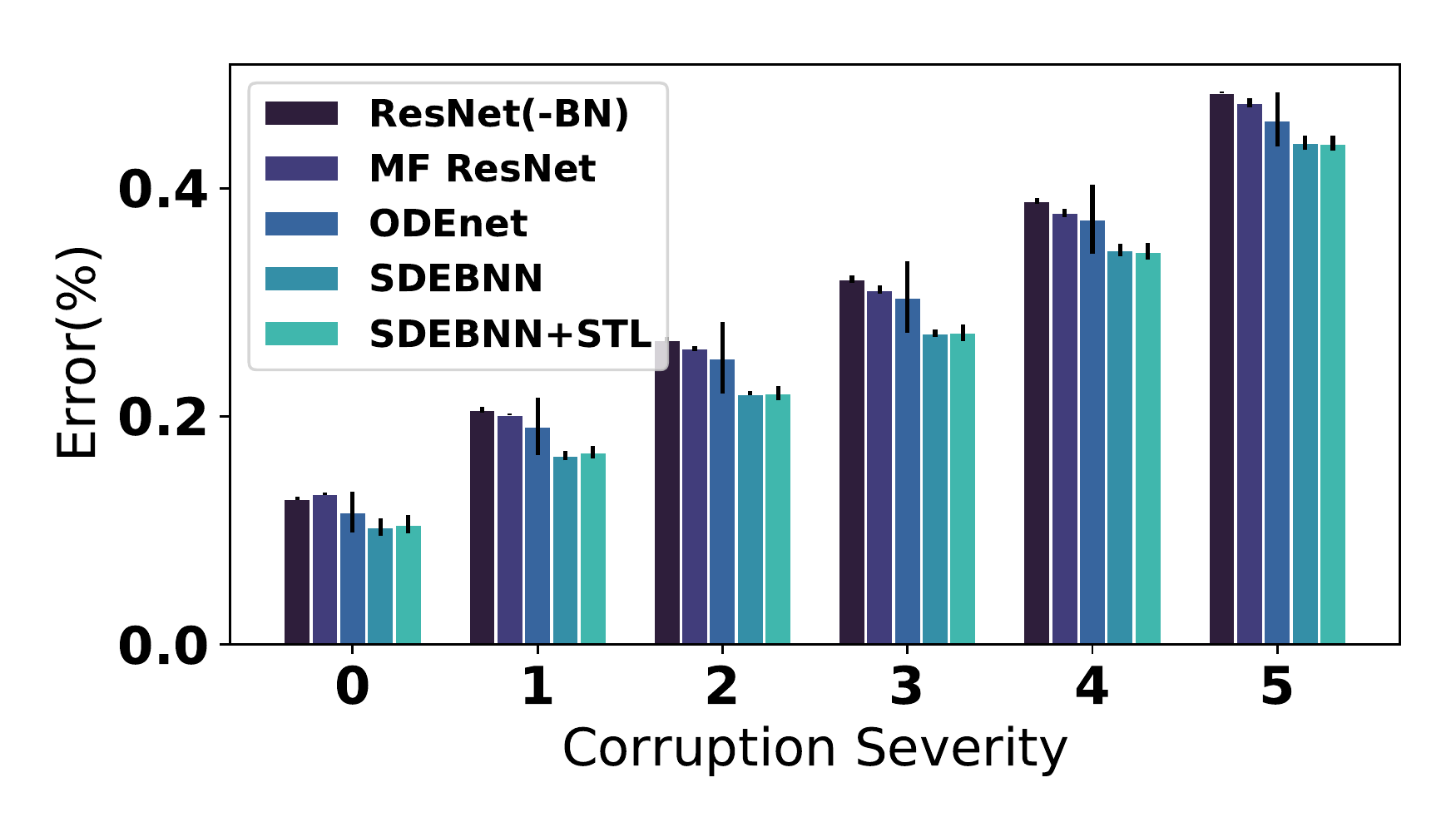}
         \label{fig:corruption_accs}
     \end{subfigure}\\
     \begin{subfigure}[b]{0.8\linewidth}
         \centering
         \includegraphics[width=\linewidth]{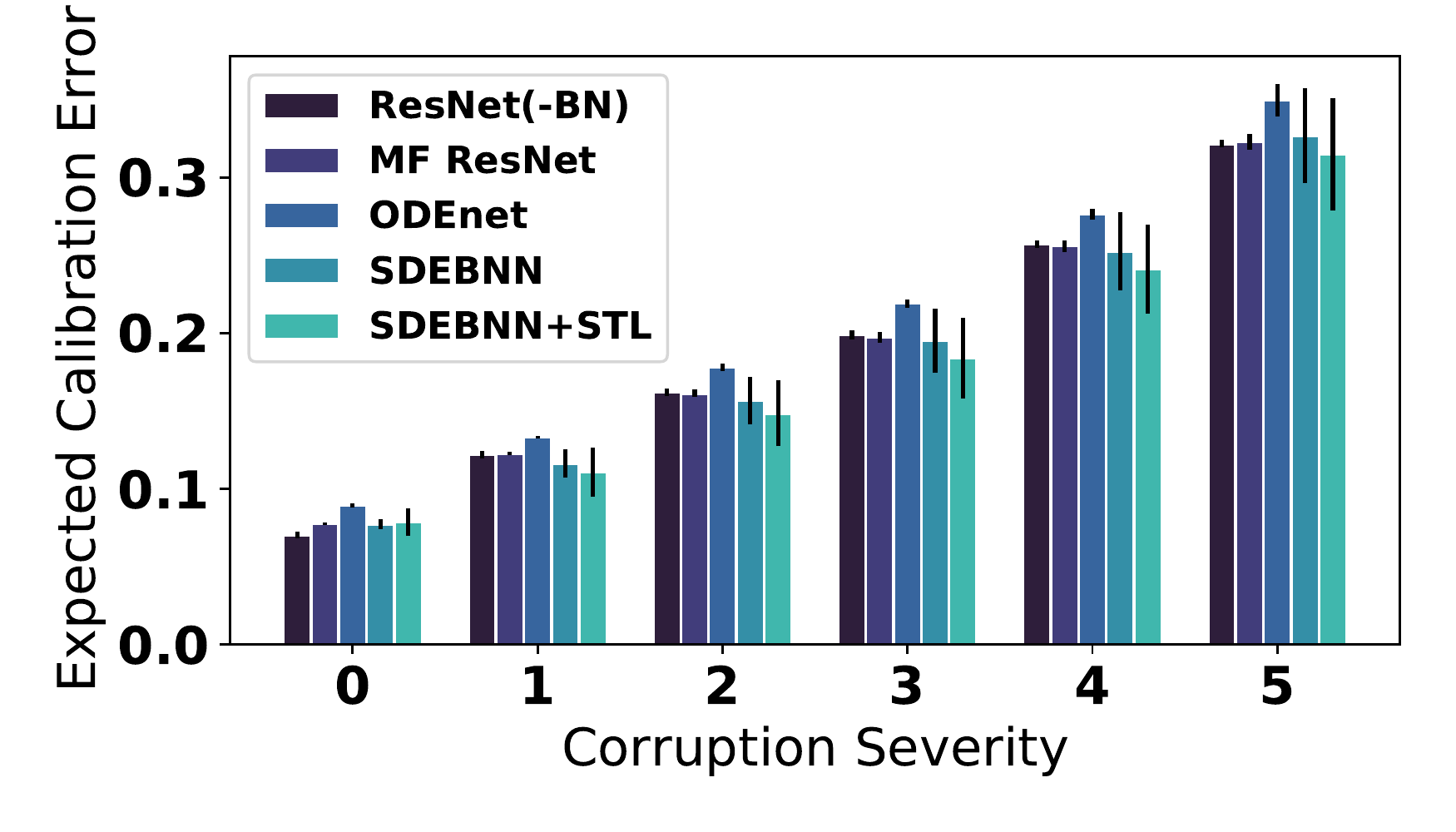}
         \label{fig:corruption_eces}
     \end{subfigure}
    \caption{CIFAR10-C. Robustness to distributional shifts on CIFAR-10. SDE-based neural nets show better accuracy and calibration than non-Bayesian and mean-field methods. Black bars show standard deviation over 3 seeds.}
    \label{fig:severitycorruption}
\end{figure}
\subsubsection{Robustness to Input Corruption}
We report the robustness of SDE-BNNs by evaluating on all 19 non-adversarial corruptions across 5 severity levels in  CIFAR10-C~\cite{hendrycks2019robustness}.
These corruptions mimic real-world perturbations such as noise, blur, and weather.
To evaluate the classification robustness of SDE-BNN, we compare the mean corruption error (mCE), an average error for each intensity level summed across all 19 perturbations, to the top-1 error rate on the corresponding clean CIFAR-10.

Figure~\ref{fig:severitycorruption} shows error on the corrupted test set relative to uncorrupted data, demonstrating a steady increase in mCE across increasing perturbation levels along with the overall error measurement summarized in Table~\ref{tab1:img_results}. On both CIFAR-10 and CIFAR10-C, the SDE-BNN and SDE-BNN + STL models achieve lower overall test error and better calibration than the baselines.

Compared to standard baselines (ResNet32 and Mean Field (MF) ResNet32), SDE-BNN achieves around 4.4\% lower absolute corruption error (CE), the total classification error for all corruption tasks across all 5 severity levels \citep{hendrycks2019robustness}, in comparison to the clean errors. The effectiveness of learned uncertainty on out-of-domain inputs indicates that SDE-BNN is more robust to observation noise despite not being trained on such diverse forms of corruptions. 

\section{SCOPE AND LIMITATIONS}\label{sec:scopelimitations}

\paragraph{Computational speed}
The cost of evaluating our model grows in $\mathcal{O}(DT)$, where $D$ is the number of weights, and $T$ the number of iterations taken by the solver.
This may seem advantageous compared to the $\mathcal{O}(D^3)$ cost for non-factorized Gaussian approximate posteriors, but the number of steps required is difficult to characterize.
Although our approach allows adjustment of the computational cost at test time, it is harder to control the cost of evaluation during training time, making our method relatively slow to train.
However, it should be straightforward to regularize these models to be faster to solve, as in \citet{kelly2020easynode}.
Relatedly, \citet{dusenberry2020efficient} recently demonstrated an $\mathcal{O}(DK)$ cost approximate posterior in standard BNNs.

\paragraph{Batch norm}
We did not incorporate batch normalization~\citep{ioffe2015batch} in any of our neural network components.
Introducing any normalization (e.g. batchnorm, layernorm, etc.) compromises the Lipschitz property required for SDEs to have a unique solution. 
Since BN introduces dependence between samples within a batch, it is also unclear how to incorporate BN while maintaining the consistency properties of Bayesian inference.
\citet{zhang2019fixup, Chang2020PrincipledWI} proposed initializations that yield the same performance without needing batch normalization.

\paragraph{Low-variance gradients for other domains}
Our extended STL gradient estimator~\citep{roeder2017sticking} to the infinite-dimensional variational objective could be applied to other settings for faster convergence, e.g. time series applications \citet{li2020scalable} investigated.

\section{RELATED WORK}\label{sec:related work}
\paragraph{Initial theoretical investigations}
The earliest theoretical treatment of infinitely-deep Bayesian neural networks was made by \citet[Chapter 2]{neal2012bayesian}, but no practical training or evaluation method was proposed.
\citet{DuvRipAdaGha2014} also investigated the theoretical properties of kernel-based constructions of infinitely-deep Bayesian neural networks.

\paragraph{Diffusion limits of discrete-time models}
We expect existing discrete-depth constructions to converge to diffusion limits in the infinitesimal limit if a system is updated with appropriately scaled Gaussian noise at each timestep. 
\citet{peluchetti2020infinitely,peluchetti2020doubly} show this holds for the output of residual networks with shallow residual blocks whose weight initializations are appropriately scaled. 
While our construction of SDE-BNN given by~\eqref{augsde} seems similar \cite{peluchetti2020infinitely}, there are two key differences: (i) We strictly enforce hidden states to follow a diffusion throughout training by directly learning a neural SDE, whereas~\citet{peluchetti2020infinitely} only ensures SDE-driven dynamics at initialization. (ii) We adopt a more general neural net architecture for the residual blocks than the shallow ones considered in~\citep{peluchetti2020infinitely}. Their work mainly discusses the convergence of shallow ResNets to SDEs, in order to analyze training stability for regular ResNets and verify that a scaled gradient formulation leads to faster convergence at the first epoch.
The consequence of (i) is that operations on diffusions (e.g., computing path-space KL) remain applicable even after our model has been trained.
While (ii) appears to be a minor difference, it actually uncovers a fundamental distinction in our analysis: Since we start out with an SDE, and only discretize for numerical computations, our model is able to incorporate any type of Lipschitz smooth residual block. Additionally, no training algorithm was specified for learning SDE models.
The analysis by \citet{peluchetti2020infinitely} relies on Taylor expanding the residual block function, which is not easy in the presence of complex residual block architectures and would require modifications to the initialization.
\citet{tzen2019neural} show that particle trajectories of the approximate posterior in discrete deep latent Gaussian models converge to a diffusion, and that the ELBO may be written with KL of measures on path space.
This construction has been explored in various forms in the past~\citep{opper2019variational, archambeau2008variational}.

\paragraph{Neural SDEs with other training objectives} Models making use of SDEs have appeared in the past, though many make use of somewhat \textit{ad-hoc} combinations of methods involving both discrete and continuous components.
\citet{kong2020sde} proposed fitting a neural SDE by using a heuristic training objective based on encouraging the diffusion to be large away from the training data and a fixed Euler-Maruyama (E-M) discretization.
\citet{innes2019zygote} trained neural SDEs by backpropagating through the operations of the solver, however their training objective simply matched the first two moments of the training data, implying that it could not consistently estimate diffusion functions.
This approach is also relatively memory-intensive.
\citet{liu2019neural} and \citet{oganesyan2020stochasticity} add noise to the solver operations in a neural ODE, although the diffusion must be tuned as a hyperparameter.
\citet{hegde2018deep} proposed a form of neural SDE using Gaussian processes to parameterize the drift and diffusion functions for a fixed E-M discretization.  However, the diffusion functions are based on an \emph{ad-hoc} construction from a Gaussian process posterior conditioned on inducing points.
\citet{ryder2018blackbox} used a Gaussian process variational posterior, effectively a continuous-time analog of a mean field approximation that may not always be expressive enough to model the true posterior.
\citet{kidger2021neural} learn neural SDEs by jointly learning a discriminator~\citep{kidger2020neural} and formalize the problem as learning generative adversarial networks. However, this would involve many more hyperparameters and require extensive tuning compared to our variational inference approach.
\paragraph{ODEnets with finite-dimensional stochasticity} Some methods based on building variational autoencoders with a neural ODE share similar training objectives, since the ELBO appears frequently in posterior inference. The Latent ODE model~\citep{rubanova2019latent} only performs inference on the distribution at an initial time of a continuous hidden state. 
\citet{de2019gru} introduced stochastic jumps at data locations, and do not perform continuous-time inference. While performing amortized inference for time series modeling, \citet{yildiz2019ode} also infer the weights of an ODE drift function.
\citet{dandekar2020bayesian} have a similar setting but for supervised learning.
\paragraph{Approximate posteriors defined as neural nets}
\citet{krueger2018bayesian} and \citet{louizos2017multiplicative} use normalizing flows to construct an unfactorized, non-Gaussian approximate posterior in BNNs.
However, normalizing flows have poor scaling with dimension and point estimates were used for most of the weights in the neural network.
Table~\ref{tab5:comparemethods} in Appendix~\ref{tab5:comparemethods} compares qualities of our approach to existing methods for stochastic variational inference in BNNs.

\section{CONCLUSION}
We developed a practical method for approximate inference in continuous-depth Bayesian neural networks. Our approach exploits a special synergy between continuous-depth models and variational inference for SDEs, providing additional benefits over standard approaches.
In particular, our method allows arbitrarily-expressive, non-factorized approximate posteriors implicitly defined through neural SDEs.
We also developed an unbiased gradient estimator for SDE variational inference whose variance approaches zero as the approximate posterior approaches the true posterior.
This combination gives our family of Bayesian continuous-depth neural networks a special property, which is that the gradients' bias and variance can be made arbitrarily small during training.
Where standard applications of MFVI on continuous-depth models perform poorly, our approach brings continuous-depth Bayesian neural networks to a comparable performance with standard Bayesian neural networks.
Furthermore, we demonstrated the ability of this continuous-depth model class to use adaptive SDE solvers.
This allows a memory-efficient training, and a fine-grained trade-off between precision and speed.


\subsubsection*{Acknowledgements}
We thank Jesse Bettencourt, Radford M. Neal, and Patrick Kidger for helpful technical discussions and revisions on earlier drafts of this work. We also thank James Bradbury for his support while implementing differential equation solvers in JAX.

\bibliographystyle{apalike}
\bibliography{references}





\input{appendix.tex}

\end{document}

%% file: mathdef.tex
\newcommand*{\bracks}[1]{\left(#1\right)}  
\newcommand*{\sbracks}[1]{\left[#1\right]}  

\newcommand{\dee}{\mathop{\mathrm{d}\!}}
\newcommand{\dt}{\,\dee t}

\newcommand{\dw}{\,\dee w}

\newcommand{\dB}{\,\dee B} 

\newcommand{\norm}[1]{\left\lVert#1\right\rVert}

\newcommand{\eq}[1]{\begin{align}#1\end{align}}
\newcommand{\eqn}[1]{\begin{align*}#1\end{align*}}

\usepackage[super]{nth}

\usepackage{nicefrac}


%% file: math_commands.tex
\usepackage{amsmath,amsfonts,bm}

\def\eqref#1{equation~\ref{#1}}

\def\1{\bm{1}}

\def\ra{{\textnormal{a}}}

\DeclareMathAlphabet{\mathsfit}{\encodingdefault}{\sfdefault}{m}{sl}
\SetMathAlphabet{\mathsfit}{bold}{\encodingdefault}{\sfdefault}{bx}{n}

\newcommand{\E}{\mathbb{E}}

\newcommand{\R}{\mathbb{R}}

\newcommand{\KL}{D_{\mathrm{KL}}}



%% file: appendix.tex
\clearpage
\appendix

\thispagestyle{empty}

\onecolumn \makesupplementtitle

\section{PROOFS}

\paragraph{Notation.} 
Denote as $\phi$ the vector of variational parameters, $f_q$ as the approximate posterior on weights, $f_p$ as the prior on weights, $f_h$ as the dynamics of hidden units, and $\sigma$ as the diffusion function. Denote the Euclidean norm of a vector $u$ by $|u|$. For function $f$ denote its Jacobian as $\nabla f$.

\subsection{Derivation of an Alternative Monte Carlo Estimator}\label{app:mcestimator}
The goal of this section is to derive a Monte Carlo estimator of the KL-divergence on path space that is similar to the \textit{fully Monte Carlo} estimator described in~\cite{roeder2017sticking}. This will serve as the basis for the subsequent heuristic derivation of the continuous-time sticking-the-landing trick.

Let $w_0$ be a fixed initial state. Let $w_1, ..., w_N$ be states at times $\Delta t, 2\Delta t, \dots, N\Delta t = T$ generated by the Euler discretization:
\eq{
\label{eq:euler_discretize_posterior}
w_{i+1} &= w_i + f_q(w_i) \Delta t + \sigma(w_i) (B_{t + \Delta t} - B_{t} ) \\
        &= w_i + f_q(w_i) \Delta t + \sigma(w_i) \Delta t^{1/2} \epsilon_{i+1}, \quad \epsilon_{i+1} \sim \mathcal{N}(0, 1).
}
where $\{B_t\}_{t\ge 0}$ is the Brownian motion.
This implies that conditional on the previous state, the current state is normally distributed: 
\eqn{
    w_{i+1} | w_i &\sim \mathcal{N} ( w_i + f_q(w_i) \Delta t, \sigma(w_i)^2 \Delta t ) .
}
Thus, the log-densities can be evaluated as
\eq{
    \label{eq:logq_step}
    \log q(w_{i+1} | w_i) &=
        -\frac{1}{2} \log ( 2\pi \sigma(w_i)^2 \Delta t ) -\frac{1}{2} \frac{
            \bracks{w_{i+1} - (w_i + f_q(w_i) \Delta t )}^2
        }{
            \sigma(w_i)^2 \Delta t
    }, \quad i = 0, \dots N - 1.
}
On the other hand, if at any time, the next state was generated from the current state based on the prior process, we would have the following log-densities:
\eq{
    \label{eq:logp_step}
    \log p(w_{i+1} | w_i) &=
        -\frac{1}{2} \log ( 2\pi \sigma(w_i)^2 \Delta t ) -\frac{1}{2} \frac{
            \bracks{w_{i+1} - (w_i + f_p(w_i) \Delta t )}^2
        }{
            \sigma(w_i)^2 \Delta t
    }, \quad i = 0, \dots N - 1.
}
Now, we substitute the form of $w_{i+1}$ based on~\eqref{eq:euler_discretize_posterior} into~\eqref{eq:logq_step} and~\eqref{eq:logp_step} and obtain
\eqn{
\log q(w_{i+1} | w_i) =& -\frac{1}{2} \log ( 2\pi \sigma(w_i)^2 \Delta t ) -\frac{1}{2} \epsilon_{i+1}^2, \\
\log p(w_{i+1} | w_i) =&
    -\frac{1}{2} \log ( 2\pi \sigma(w_i)^2 \Delta t )
    \\&
    -\frac{1}{2} \Bigg(
        \frac{ (f_q(w_i) - f_p(w_i))^2 }{ \sigma(w_i)^2 } \Delta t + \frac{ 2 ( f_q(w_i) - f_p(w_i) ) \epsilon_{i+1} }{ \sigma(w_{i}) } \Delta t^{1/2}
        + \epsilon_{i+1}^2
    \Bigg).
}
The KL divergence on the path space could then be regarded as a sum of infinitely many KL-divergences between Gaussians:
\eq{
     &\lim_{N \to \infty} \sum_{i=0}^N \E_{w_i} \sbracks{ \KL \bracks{q(w_{i+1} | w_i) || p(w_{i+1} | w_i)}  } \\
    =&
        \lim_{N \to \infty} \sum_{i=0}^N \E_{w_i} \sbracks{
        \E_{w_{i + 1} \sim q(w_{i+1} | w_i)} \sbracks{
        \log \frac{q(w_{i+1} | w_i)}{p(w_{i+1} | w_i)}
        }
    } \\
    =&
        \lim_{N \to \infty} \sum_{i=0}^N \E_{w_i} \sbracks{
            \E_{\epsilon_{i+1}} \sbracks{
                \frac{ (f_q(w_i) - f_p(w_i))^2 }{ 2 \sigma(w_i)^2 } \Delta t + 
                \frac{ (f_q(w_i) - f_p(w_i)) }{ \sigma(w_i) } \Delta t^{1/2} \epsilon_{i+1}
            }
        } \\
    =& \;
        \E \sbracks{
            \frac{1}{2} \int_0^T | u_t |^2 \dt + \int_0^T u_t \dB_t
        }.
    \label{eq:fmc}
}

\subsection{Sticking-the-landing in Continuous Time}
\label{app:stlheuristic}
For a non-sequential latent variable model, the sticking-the-landing (STL) trick removes from the fully Monte Carlo ELBO estimator a score function term of the form $\partial \log q(w, \phi) / \partial \phi$, where $w$ is sampled using the reparameterization trick and may depend on $\phi$. 
The score function term has $0$ expectation, but may affect the variance of the gradient estimator for the inference distribution's parameters.

Here, we exploit this intuition and apply it to each step before taking the limit. More precisely, we apply the STL trick to estimate the gradient of $\KL ( q(w_{i+1} | w_i) || p(w_{i+1} | w_i) )$ for $i=1, 2, \dots ,N$, and thereafter take the limit as the mesh size of the discretization goes to $0$. For each individual term, the score function term to be removed is 
\eqn{
    \frac{\partial }{\partial \phi} \log q(w_{i+1} | w_i, \phi) =&
        -\frac{1}{ 2 \sigma^2(w_i) \Delta t } \frac{\partial }{\partial \phi}
        \sbracks{
            \bracks{
                w_{i+1} - (w_i + f_q( w_i, \phi) \Delta t)
            }^2
        } \\
    =&
        \frac{\partial }{\partial \phi} \sbracks{
            \frac{ f_q(w_i, \phi) }{ \sigma(w_i) }
        } \epsilon_{i+1} \Delta t^{1/2}.
}
Now, we sum up all of these terms and take the limit as $\Delta t \to 0$. This gives us
\eqn{
    &\lim_{N \to \infty} \sum_{i=0}^N \E_{w_i} \sbracks{
        \E_{ w_{i+1} \sim q(w_{i+1} | w_i) } \sbracks{
            \frac{\partial }{\partial \phi} \log q(w_{i+1} | w_i)
        }
    }\\
    =&
        \lim_{N \to \infty} \sum_{i=0}^N \E_{w_i} \sbracks{
            \E_{ \epsilon_{i+1} } \sbracks{
                \frac{\partial }{\partial \phi} \sbracks{
                    \frac{ f_q(w_i, \phi) }{ \sigma(w_i) }
                } \epsilon_{i+1} \Delta t^{1/2}
            }
        } \\
    =&
        \; \E\sbracks{
            \int_0^T \frac{\partial }{\partial \phi} \sbracks{ 
                \frac{ f_q(w_t, \phi) }{ \sigma(w_t) }
            } \dB_t
        } \\
    =&
        \; \E\sbracks{
            \int_0^T
            \frac{\partial }{\partial \phi} \sbracks{u_t} \dB_t
        }.
}


Removing this term from the fully Monte Carlo estimator in~\eqref{eq:fmc} gives rise to the following estimator of a surrogate objective that facilitates implementation:
\eqn{
     \widehat{\text{ELBO}} =&
        \log p ( \mathcal{D} \mid  w ) 
        - \int_{t_0}^{t_1} \frac{1}{2} \norm{ u(w_t, t, \phi) }_2^2 \dt 
        \\&
        - \int_{t_0}^{t_1} u(w_t, t, \texttt{stop\_gradient}(\phi)) \dB_t, \quad w(\cdot) \sim q_\phi().
}


\newpage
\section{EXPERIMENTAL SETTINGS}
\begin{table*}[!htbp]
\centering
\setlength{\tabcolsep}{0.8em}
\resizebox{1\textwidth}{!}{
\begin{tabular}{@{} l l l l l l @{}}\toprule
  & \multicolumn{1}{c}{\textbf{\textsc{}}} &
  & \multicolumn{1}{c}{\textbf{\textsc{Experiments}}} &
  & \multicolumn{1}{c}{\textbf{\textsc{}}} \\
\cmidrule(l){3-5} 
Model & Hyper-parameter & 1D Regression
& MNIST~\cite{deng2012mnist} & CIFAR-10~\cite{krizhevsky2014cifar} \\
\cmidrule(r){1-2} \cmidrule(l){3-5} 
ResNet32 
 & Learning Rate & -- & 1e-3 & 7e-4 \\
 & Batch Size & -- & 128 & 128 \\
 & Activation & -- & tanh & tanh \\
 & Epochs & -- & 100 & 500 \\
\cmidrule(r){1-2} \cmidrule(l){3-5} 
ODEnet
 & Augment dim. & 2 & 2 & 2 \\
 & \# blocks & 1 & 1 & 2-2-2 \\
 & Diffusion $\sigma$ & 0 & 0 & 0 \\
 & KL coef. & 0 & 0 & 0 \\
 & Learning Rate & 1e-3 & 1e-3 & 7e-4 \\
 & \# Solver Steps & 10 & 20 & 20 \\
 & Batch Size & 40 & 128 & 128 \\
 & Activation & swish & tanh & tanh \\
 & Epochs & 800 & 100 & 500 \\
\cmidrule(r){1-2} \cmidrule(l){3-5} 
HyperODEnet 
 & <ODEnet> & -- & <ODEnet> & <ODEnet> \\
 & KL coef. & -- & 1e-3 & 1e-3 \\
 & Drift $f_w$ dim. & -- & 1-64-1 & 1-128-1 \\
\cmidrule(r){1-2} \cmidrule(l){3-5} 
MFVI ResNet32 
 & <ResNet32> & -- & <ResNet32> & <ResNet32> \\
 & KL coef. & -- & 1e-3 & 1e-3 \\
\cmidrule(r){1-2} \cmidrule(lr){3-5}
MFVI ODEnet 
 & <ODEnet> & -- & <ODEnet> & <ODEnet> \\
 & KL coef. & -- & 1e-3 & 1e-3 \\
\cmidrule(r){1-2} \cmidrule(lr){3-5}
MFVI HyperODEnet 
 & <MFVI ODEnet> & -- & <MFVI ODEnet> & <MFVI ODEnet> \\
 & Drift $f_w$ dim. & -- & 1-64-1 & 1-128-1 \\
\cmidrule(r){1-2} \cmidrule(lr){3-5}
SDE BNN 
 & <ODEnet> & <ODEnet> & <ODEnet> & <ODEnet> \\
 & Learning Rate & 1e-3 & 1e-3 & 7e-4 \\
 & \# blocks & 1 & 1 & 2-2-2 \\
 & Drift $f_x$ dim. & 32 & 32 & 64 \\
 & Drift $f_w$ dim. & 32 & 1-64-1 & 2-128-2 \\
 & Diffusion $\sigma$ & 0.2 & 0.1 & 0.1 \\
 & \# Posterior Samples & 20 & 1 & 1 \\
\cmidrule(r){1-2} \cmidrule(lr){3-5}
SDE BNN (+ STL) & <SDE BNN> & <SDE BNN> & <SDE BNN> & <SDE BNN> \\
\bottomrule
\end{tabular}
}
\caption{These are the hyper-parameters for each method of evaluation pertaining to results in the toy and classification tasks of Table \ref{tab1:img_results}. Each model was run on a single Nvidia RTX6000 GPU on our compute clusters. SDE and learning optimization parameters were tuned according to a validation set sampled randomly from 10\% of the training set. No schedules of any kind on the hyper-parameters were used in training. Settings with high overlap with another model are indicated using \texttt{<model>} with additional parameters overridden as necessary. Each block is separated either by a downsampling or upsampling convolutional layer (i.e., the \texttt{-}'s).}
\label{tab:hparam settings}
\end{table*}

\newpage
\section{ADDITIONAL RESULTS}\label{app:additional_figures}

\subsection{Augmentation in Differential Equation Models}

\begin{figure}[h]
    \centering
    \begin{subfigure}[b]{0.32\linewidth}
         \centering
         \includegraphics[trim=8cm 0 0 0,clip,width=\linewidth]{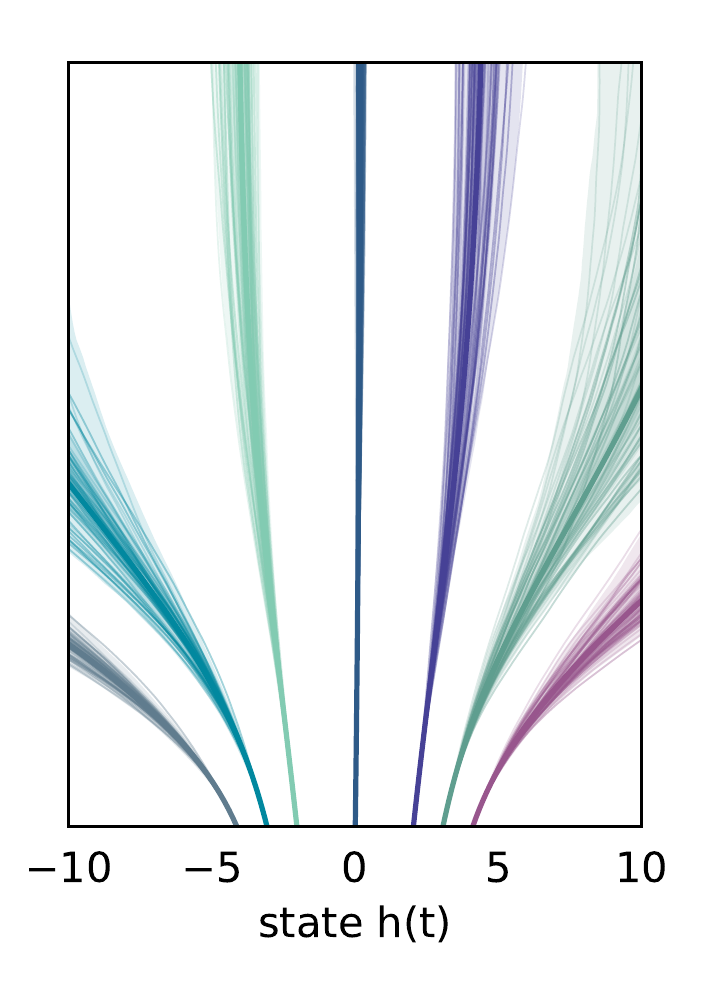}
         \caption{Non-augmented dimension}
    \end{subfigure}
    \begin{subfigure}[b]{0.32\linewidth}
         \centering
         \includegraphics[trim=8cm 0 0 0,clip,width=\linewidth]{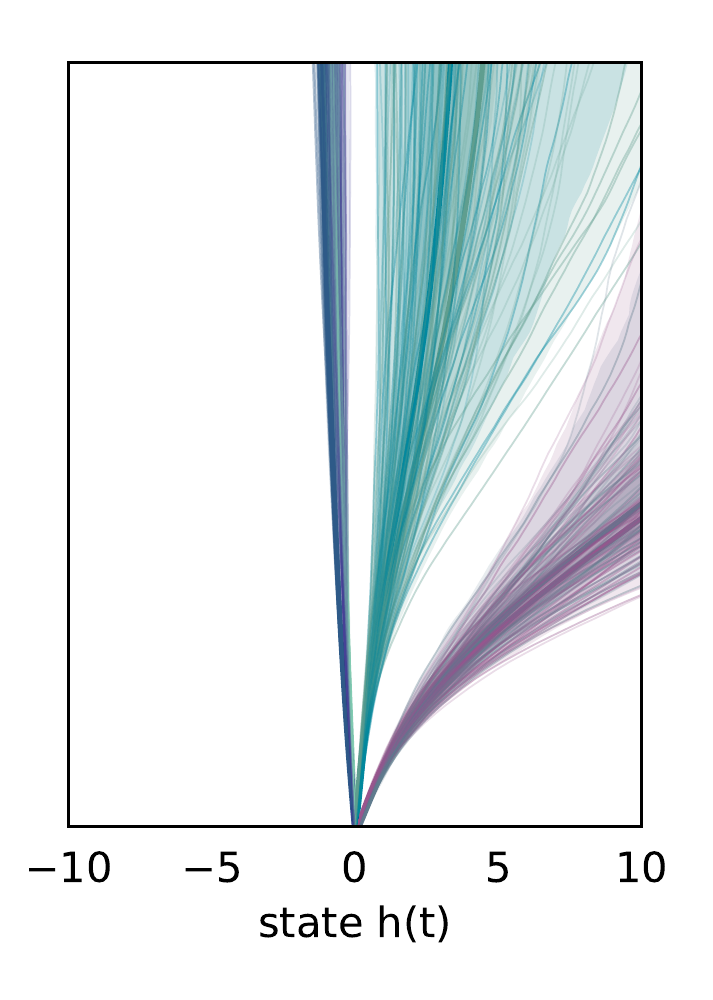}
         \caption{from 2nd augmented dimension}
    \end{subfigure}
    \begin{subfigure}[b]{0.32\linewidth}
         \centering
         \includegraphics[trim=8cm 0 0 0,clip,width=\linewidth]{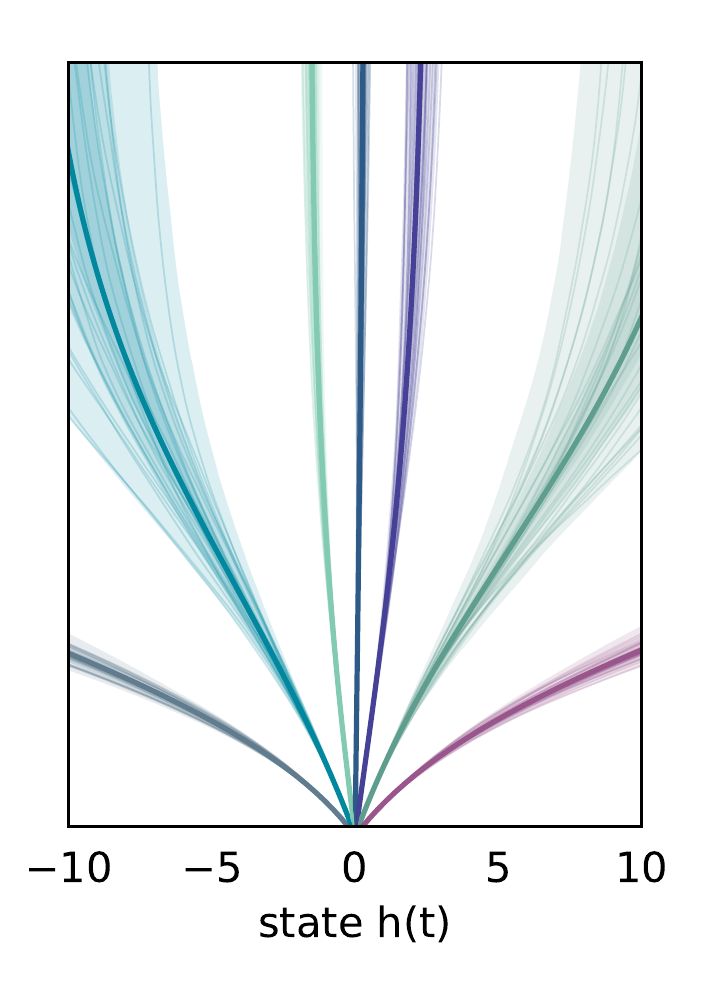}
         \caption{from last augmented dimension}
     \end{subfigure}
     \vspace{-0.6em}
    \caption{Example flows sampled from learned SDE dynamics. All continuous-depth models were trained by augmenting the state by 2 dimensions, refer to Figure~\ref{fig:dynamics} for main results. \textit{Left:} The SDE-BNN learns meaningful parameterizations on the non-extraneous dimensions of the input state vector. In the case of a true function being monotonic, the augmented dimensions simply help the main output.
    \textit{Middle:} The model learns to ignore dimensions that are not necessary to train on, especially on simpler tasks as in the toy setting. Samples in augmented dimensions can overlap for different input values in the given domain $(-5, 5)$. \textit{Right:} Similarly, the last output dimension was also associated with augmentation and was not a well learned representation of the data, ignoring the initial inputs entirely (all values are 0).}
    \label{fig:posteriorpredictive}
\end{figure}

\subsection{Classification Results}
\begin{table*}[!htbp]
\centering
\setlength{\tabcolsep}{0.8em}
\resizebox{1\textwidth}{!}{
\begin{tabular}{@{} l c c c c c c @{}}\toprule
  & \multicolumn{2}{c}{\textbf{\textsc{MNIST}}} &
  & \multicolumn{2}{c}{\textbf{\textsc{CIFAR-10}}} \\
\cmidrule(lr){2-3} \cmidrule(l){5-6} 
Model & \tcen{Accuracy (\%) } & \tcen{ECE ($\times 10^{-2}$)}
& & \tcen{Accuracy (\%)} & \tcen{ECE ($\times 10^{-2}$)} \\
\cmidrule(r){1-1}\cmidrule(lr){2-3} \cmidrule(l){5-6} 
SDE BNN & 99.30 $\pm$ 0.09 & 0.63 $\pm$ 0.10 & &  88.08 $\pm$ 1.25 & 7.53 $\pm$ 0.44 \\
SDE BNN (+ STL) & 99.10 $\pm$ 0.09 & 0.78 $\pm$ 0.12 & & 87.95 $\pm$ 1.32 & 7.94 $\pm$ 0.59 \\
SDE BNN $w0$ inferred & 99.04 $\pm$ 0.03 & 0.73 $\pm$ 0.04 & & 88.04 $\pm$ 0.30 &  6.56 $\pm$ 0.62 \\
SDE BNN $w0$ inferred (+ STL) & 99.05 $\pm$ 0.00 & 0.79 $\pm$ 0.04 & &  87.37 $\pm$ 0.63  &  6.44 $\pm$ 0.10\\
\bottomrule
\end{tabular}
}
\caption{Classification accuracy and expected calibration error on MNIST ($100^{th}$ epoch) and CIFAR-10 ($300^{th}$ epoch) for additional baseline to the ones in Table~\ref{tab1:img_results}. Values are compared at the 100th epoch for MNIST and 300th for CIFAR-10. Here $w0$, the initial drift of the posterior SDE, is inferred using a Gaussian prior rather than being a fixed value. The best prior variance was selected in a preliminary sweep between values in the range [0.1, 0.44]. The performance is slightly worse than the point estimate but displays better calibration as a trade-off. It can be noted that calibration may appear better earlier on in training, as in prior to converging and reaching non-uniform confidence, but model predictions are not necessarily correct.}
\label{tab3:add_img_results}
\end{table*}
 
\newpage
\subsection{Sticking the Landing Results}
\begin{table*}[!htbp] 
\centering
\setlength{\tabcolsep}{1em}
\resizebox{1\textwidth}{!}{
\begin{tabular}{lccc}
\hline
Method &
\begin{tabular}[c]{@{}c@{}}Accuracy (\%)\end{tabular} &
\begin{tabular}[c]{@{}c@{}}Negative \\ Log-likelihood ($\times 10^{-4}$) \end{tabular} &
\begin{tabular}[c]{@{}c@{}}ELBO \end{tabular} \\ 
\hline
\begin{tabular}[c]{@{}l@{}}SDE BNN\end{tabular} & 95.91 $\pm$ 0.2 & 1.17 $\pm$ 0.309 & 1.40 $\pm$ 0.2 \\ \hline
SDE BNN (+STL) & 96.89 $\pm$ 0.2 & 0.309 $\pm$ 0.15 & 1.183 $\pm$ 0.2  \\
\end{tabular}
}
\caption{Training with STL estimator on CIFAR-10 shows training time improvements in accuracy, negative log likelihood, and ELBO objective in addition to reducing variance. This improvement to the standard gradient estimator can be especially useful in settings where the approximate posterior is sufficiently flexible (i.e. the drift neural net is very large relative to the state size).}
\label{tab4:stl_improvement}
\end{table*}

\subsection{Calibration Results}
\label{app:calibration}
\begin{figure}[H]
    \centering
    \begin{subfigure}[b]{0.45\linewidth}
         \centering
         \includegraphics[width=\linewidth]{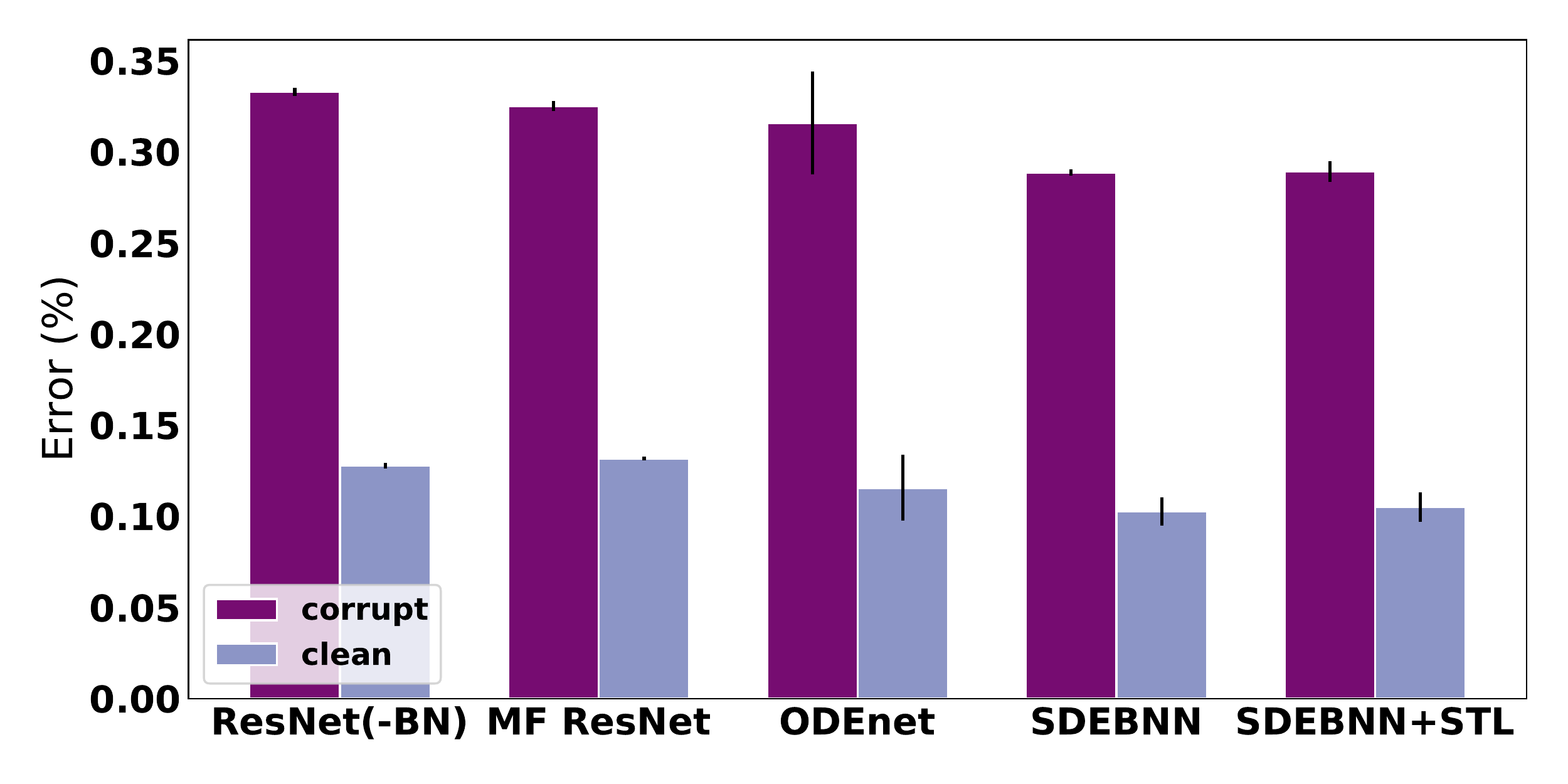}
         \caption{Model vs Classification Error}
         \label{fig:cifarcmodelvserror}
    \end{subfigure}
    \begin{subfigure}[b]{0.45\linewidth}
         \centering
         \includegraphics[width=\linewidth]{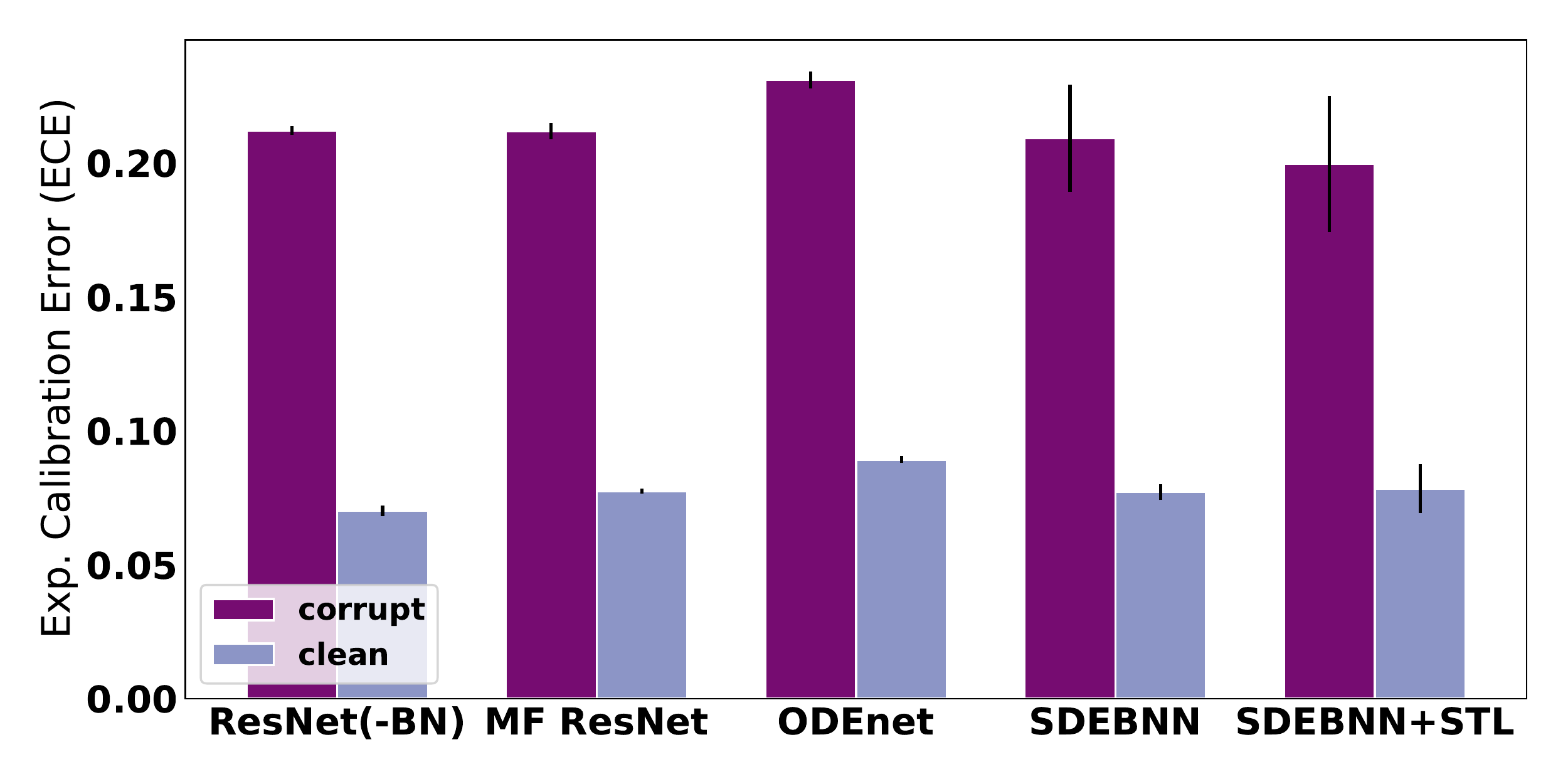}
         \caption{Model vs Expected Calibration Error}
         \label{fig:cifarcmodelvsece}
    \end{subfigure}
    \begin{subfigure}[b]{0.45\linewidth}
         \centering
         \includegraphics[width=\linewidth]{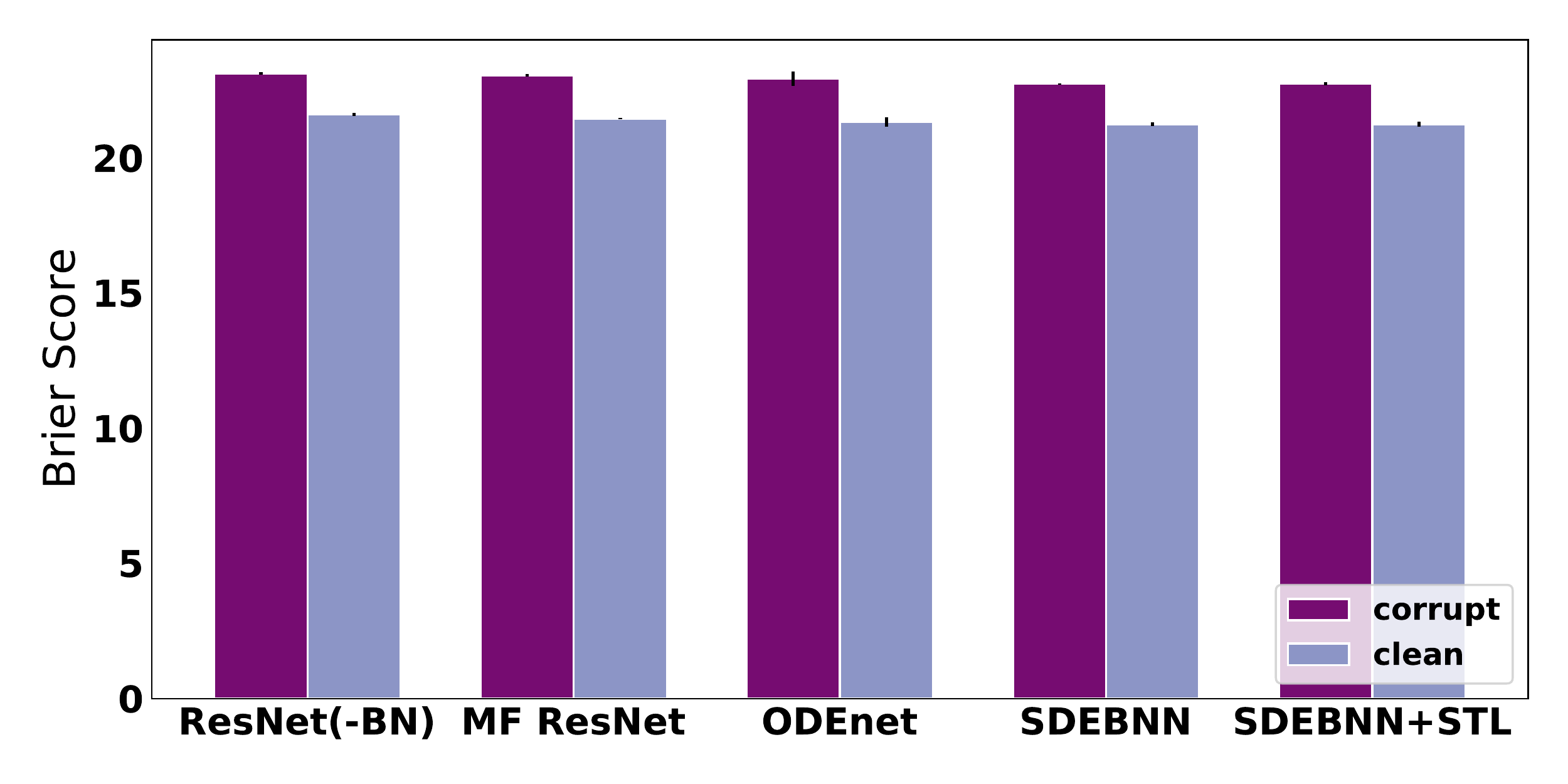}
         \caption{Model vs Brier Score}
         \label{fig:cifarcmodelvsbrier}
     \end{subfigure}
    \begin{subfigure}[b]{0.45\linewidth}
         \centering
         \includegraphics[width=\linewidth]{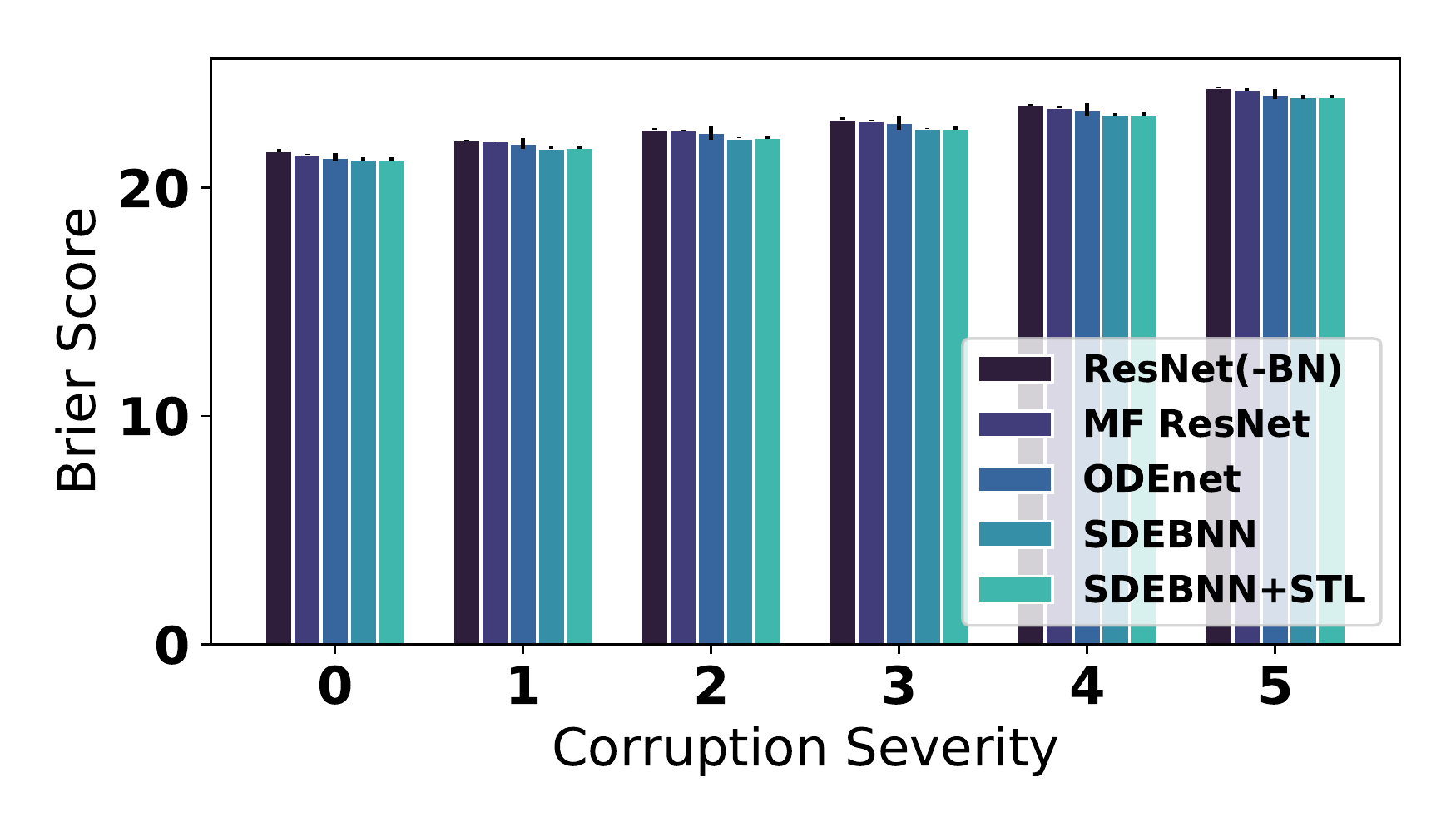}
         \caption{Corruption severity vs Brier Score}
         \label{fig:cifarcseverityvsbrier}
    \end{subfigure}
    \caption{Figures~\ref{fig:cifarcmodelvserror}-\ref{fig:cifarcmodelvsbrier} show that the SDE BNN and SDE BNN + STL models outperform their non-continuous depth ResNet counterparts on all three robustness metrics when evaluated on the corrupt CIFAR-10C benchmarks. Figure~\ref{fig:cifarcseverityvsbrier} indicates that the accuracy of predictions is relatively consistent across all severity levels with the SDE-BNN and SDE-BNN + STL models having relatively better calibrated predictions.}
    \label{fig:otherbnnevals}
\end{figure}

\newpage
\subsection{Comparisons with Other Bayesian Models}
\begin{table*}[!htbp] 
\centering
\setlength{\tabcolsep}{0.2em}
\resizebox{1\textwidth}{!}{
\begin{tabular}{lcccc}
\hline
Method &
\begin{tabular}[c]{@{}c@{}}Posterior over\\Stochastic Process\end{tabular} &
\begin{tabular}[c]{@{}c@{}}Flexible \\Approximate Posterior\end{tabular} &
\begin{tabular}[c]{@{}c@{}}Adaptive \\Computation \end{tabular} &
References \\ \hline
\begin{tabular}[c]{@{}l@{}}Bayes by Backprop\end{tabular} & \xmark & \xmark & \xmark &  \citet{blundell2015weight}  \\ \hline
MCMC for BNNs & \xmark & \cmark & \xmark & \citep{neal2012bayesian, wenzel2020good, izmailov2021bayesian}   \\
\begin{tabular}[c]{@{}l@{}}Bayesian Hypernets\end{tabular} & \xmark & \cmark & \xmark & \citet{krueger2018bayesian} \\ \hline
BBVI for SDEs & \cmark & \xmark & \xmark & \citet{ryder2018blackbox}  \\ \hline
Bayesian Neural ODEs & \xmark & \xmark & \cmark & 
\begin{tabular}[c]{@{}c@{}}\citet{yildiz2019ode}\\\citet{dandekar2020bayesian}\end{tabular} \\ \hline
SDE-BNN & \cmark & \cmark & \cmark  & current work
\end{tabular}
}
\caption{Properties of various Bayesian supervised learning approaches.}
\label{tab5:comparemethods}
\end{table*}

\subsection{Other Bayesian Methods}
\begin{figure}[!htbp] 
\vspace{-1em}
    \centering
    \subfloat{{\includegraphics[width=0.31\linewidth]{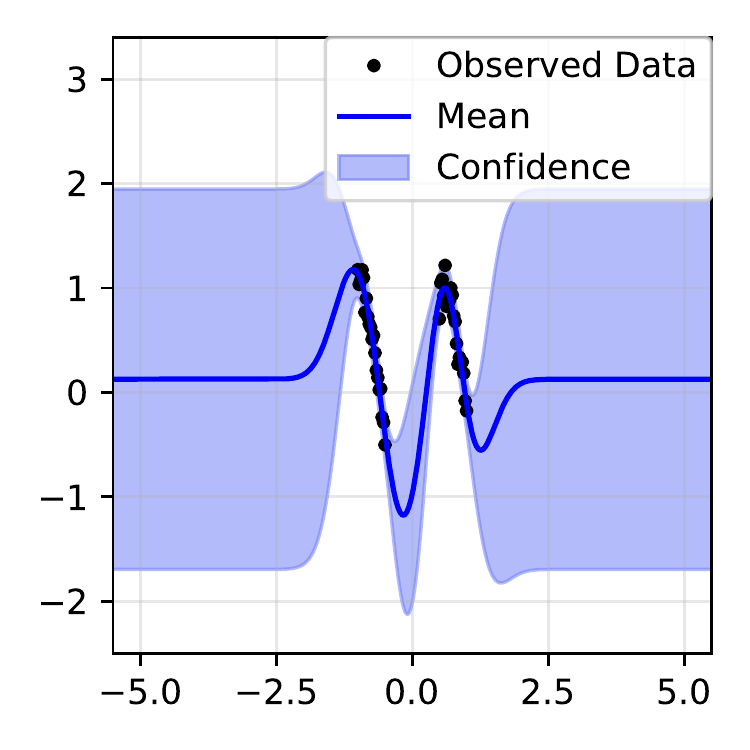} }}%
    \vspace{-0.1cm}
    \subfloat{{\includegraphics[width=0.31\linewidth]{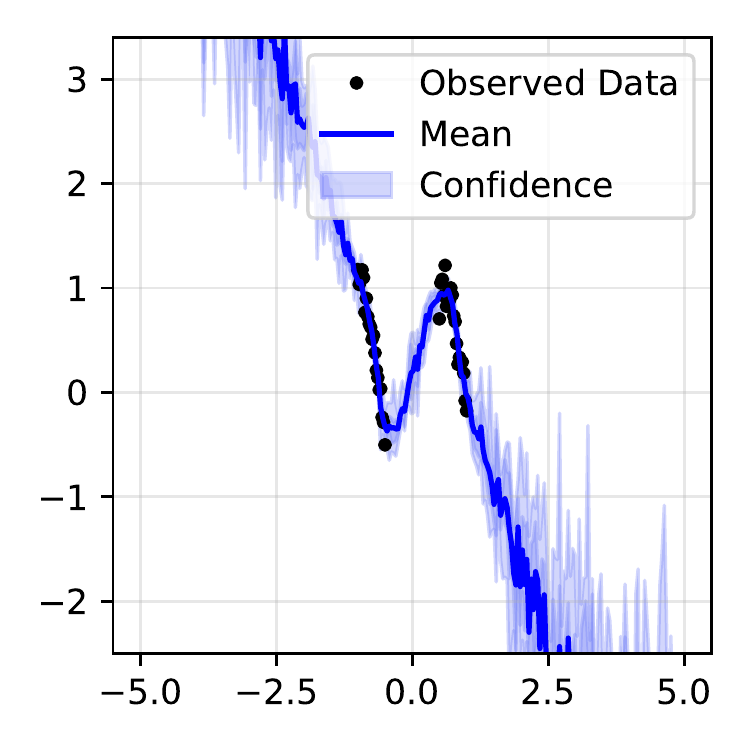} }}%
    \vspace{-0.1cm}
    \subfloat{{\includegraphics[width=0.31\linewidth]{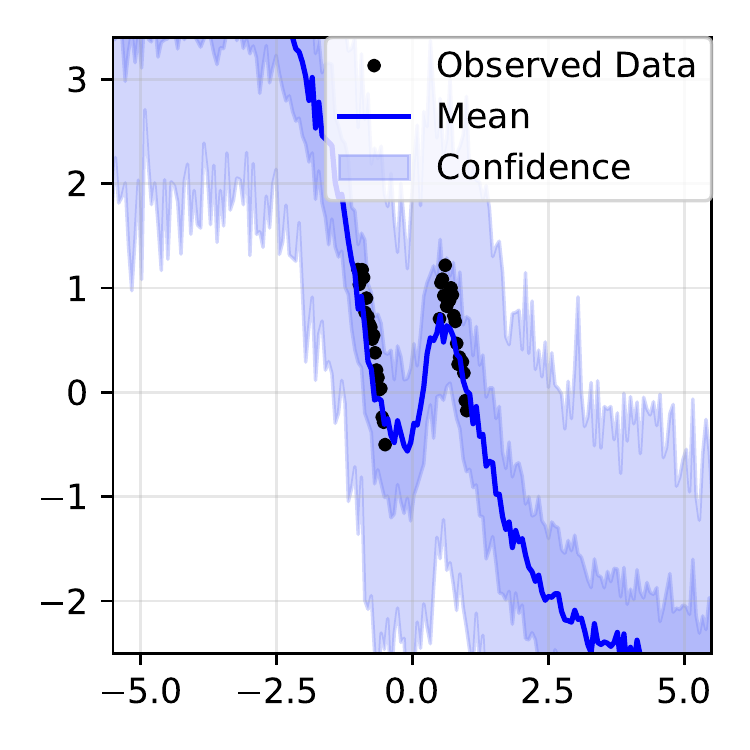} }}%
    \vspace{-0.5em}
    \caption{Approximate posteriors from other common Bayesian statistical models. \textit{Left:} Gaussian Process. \text{Center:} Deep Ensemble K=8. \text{Right:} MFVI. Different variances and extrapolations are learned from the SDE-BNN across other Bayesian model parameterizations, which can result in more or less reasonable uncertainty bounds depending on interpretation.}%
    \label{fig:morebaselines}%
\end{figure}

\subsection{Robustness to solver error at test time}
\label{app:solvererror}
\begin{figure}[H]
    \centering
    \begin{subfigure}[b]{0.23\linewidth}
    \includegraphics[width=\linewidth]{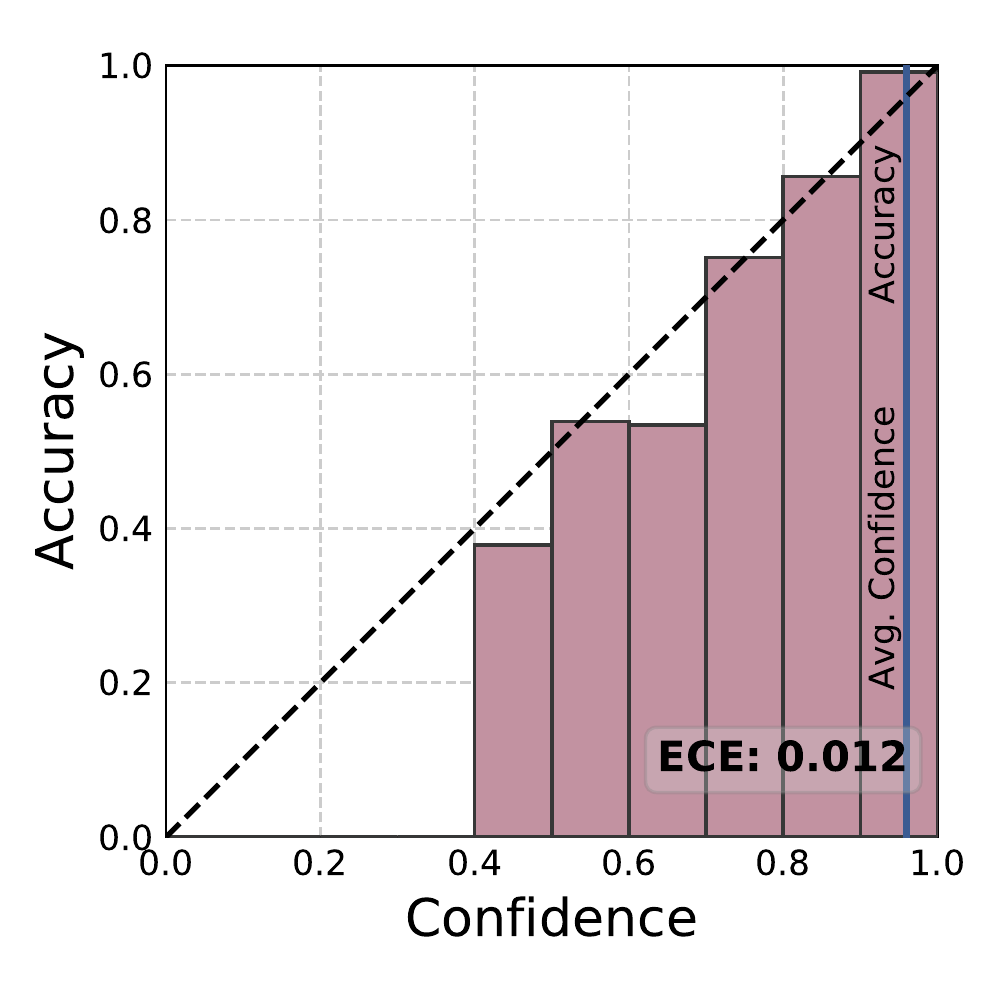}
    \vspace{-2em}
    \caption{160 steps}
    \end{subfigure}
    \begin{subfigure}[b]{0.23\linewidth}
    \includegraphics[width=\linewidth]{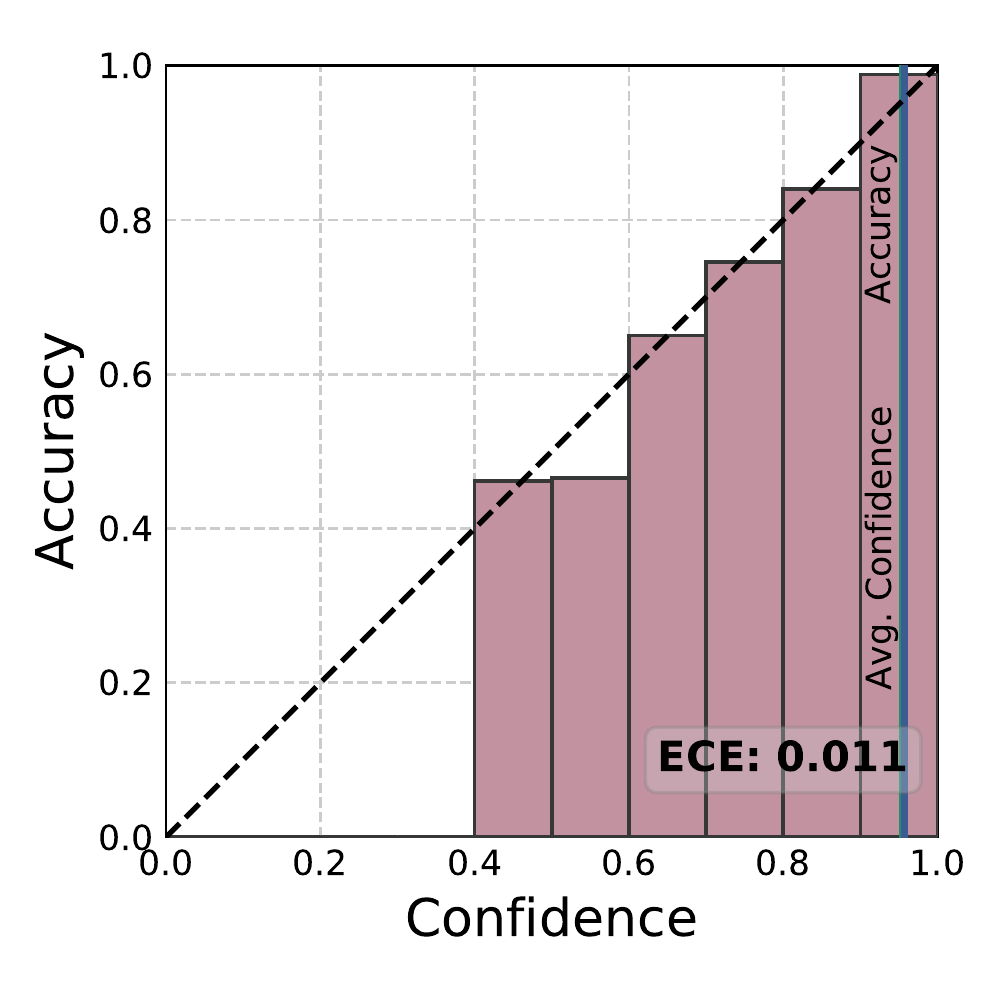}
    \vspace{-2em}
    \caption{176 steps}
    \end{subfigure}
    \begin{subfigure}[b]{0.23\linewidth}
    \includegraphics[width=\linewidth]{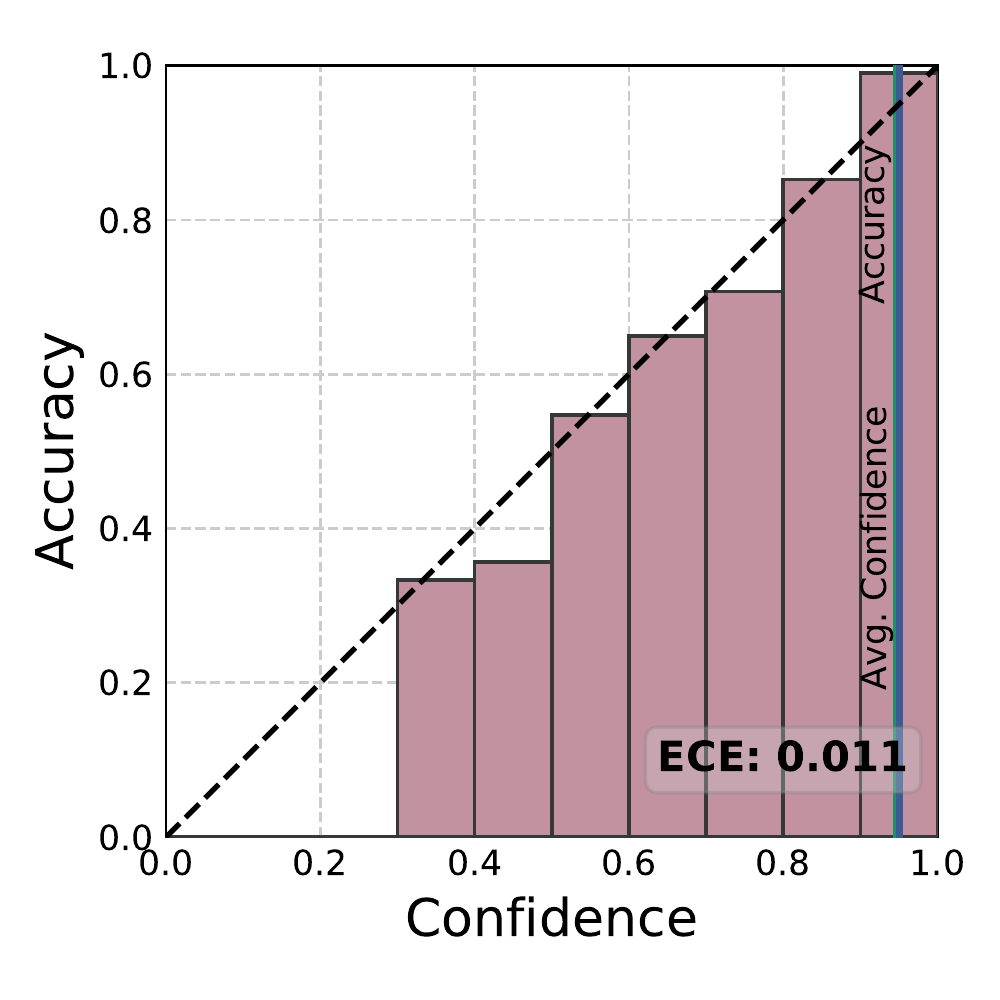}
    \vspace{-2em}
    \caption{192 steps}
    \end{subfigure}
    \begin{subfigure}[b]{0.23\linewidth}
    \includegraphics[width=\linewidth]{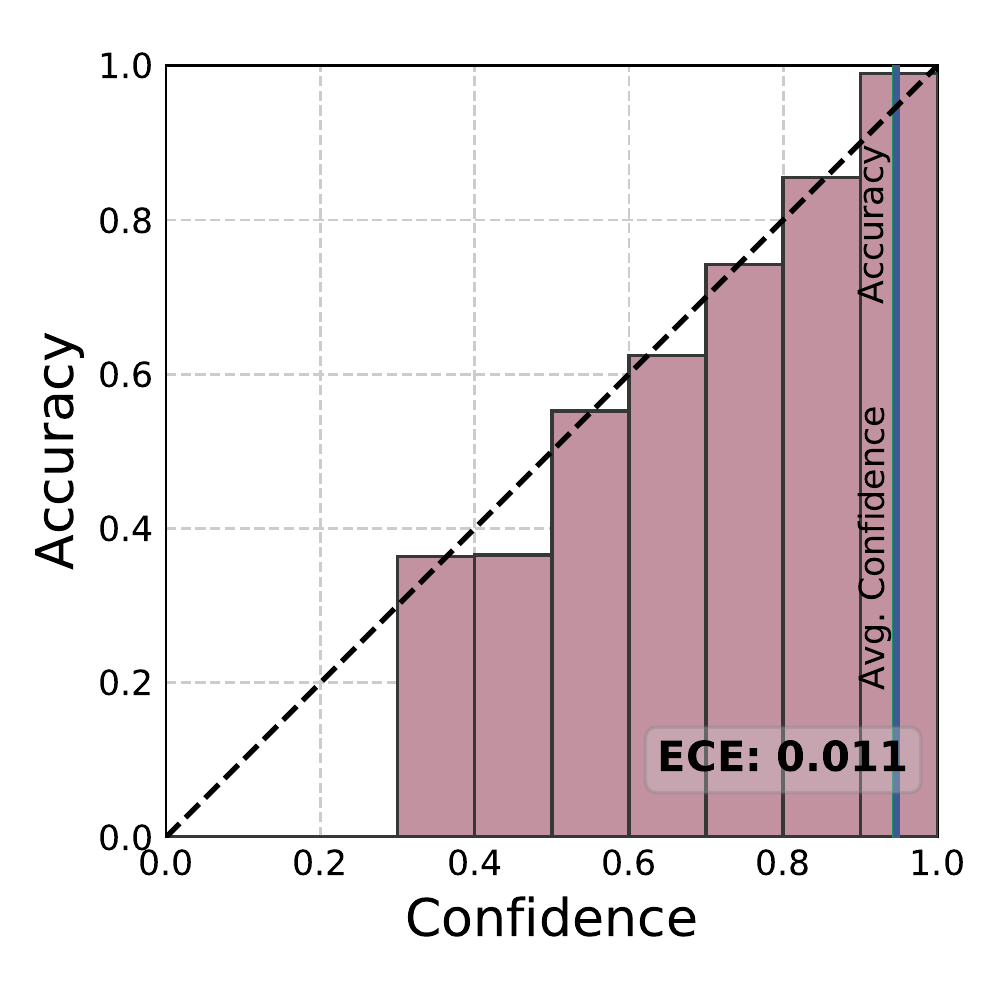}
    \vspace{-2em}
    \caption{208 steps}
    \end{subfigure}
    \caption{CIFAR10 image classification with a SDE-BNN. Better calibration can be obtained by increasing solver step sizes during inference without substantially changing the training error.}
    \label{fig:cifar10_calib_spectrum}
\end{figure}

\begin{figure}[H]
    \centering
    \begin{subfigure}[b]{0.23\linewidth}
    \includegraphics[width=\linewidth, trim=11.5px 10px 12px 19px, clip]{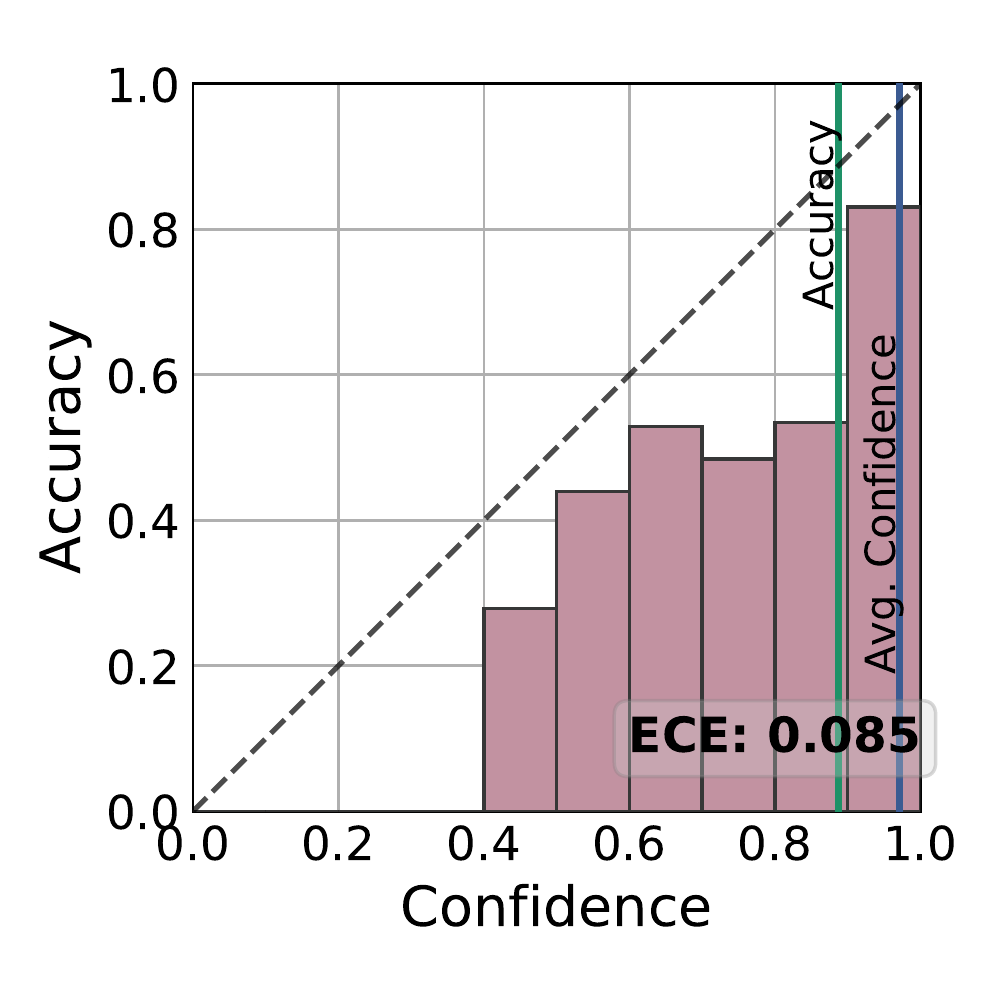}
    \vspace{-2em}
    \caption{154 steps}
    \end{subfigure}
    \begin{subfigure}[b]{0.23\linewidth}
    \includegraphics[width=\linewidth, trim=11.5px 10px 12px 19px, clip]{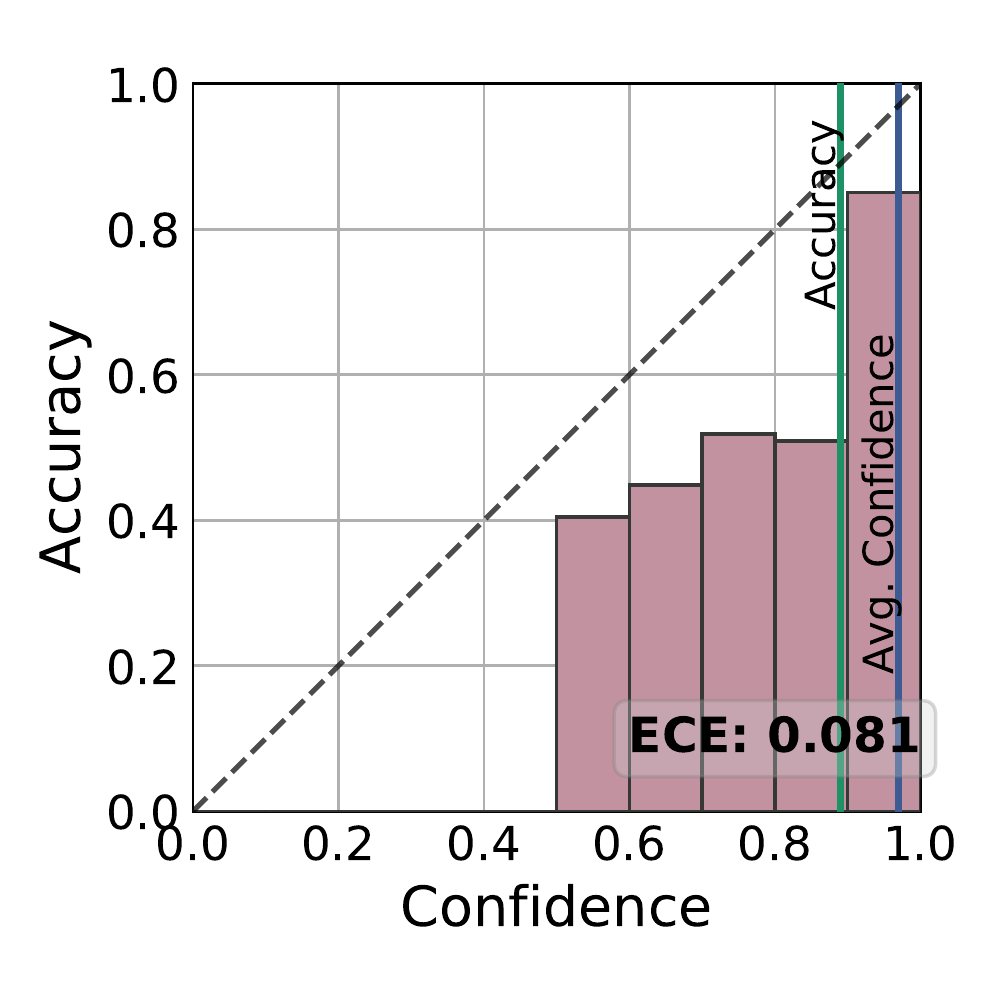}
    \vspace{-2em}
    \caption{160 steps}
    \label{fig:stepstrained}
    \end{subfigure}
    \begin{subfigure}[b]{0.23\linewidth}
    \includegraphics[width=\linewidth, trim=11.5px 10px 12px 19px, clip]{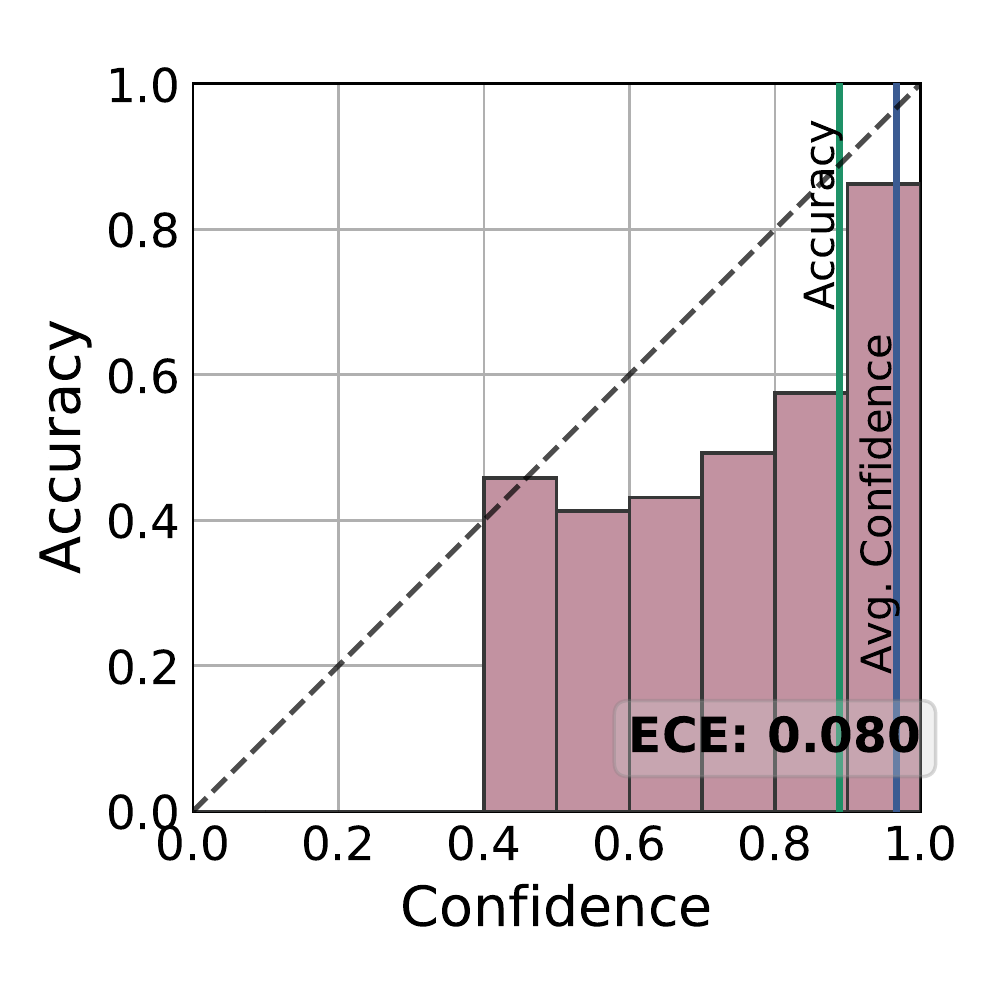}
    \vspace{-2em}
    \caption{176 steps}
    \end{subfigure}
    \begin{subfigure}[b]{0.23\linewidth}
    \includegraphics[width=\linewidth, trim=11.5px 10px 12px 19px, clip]{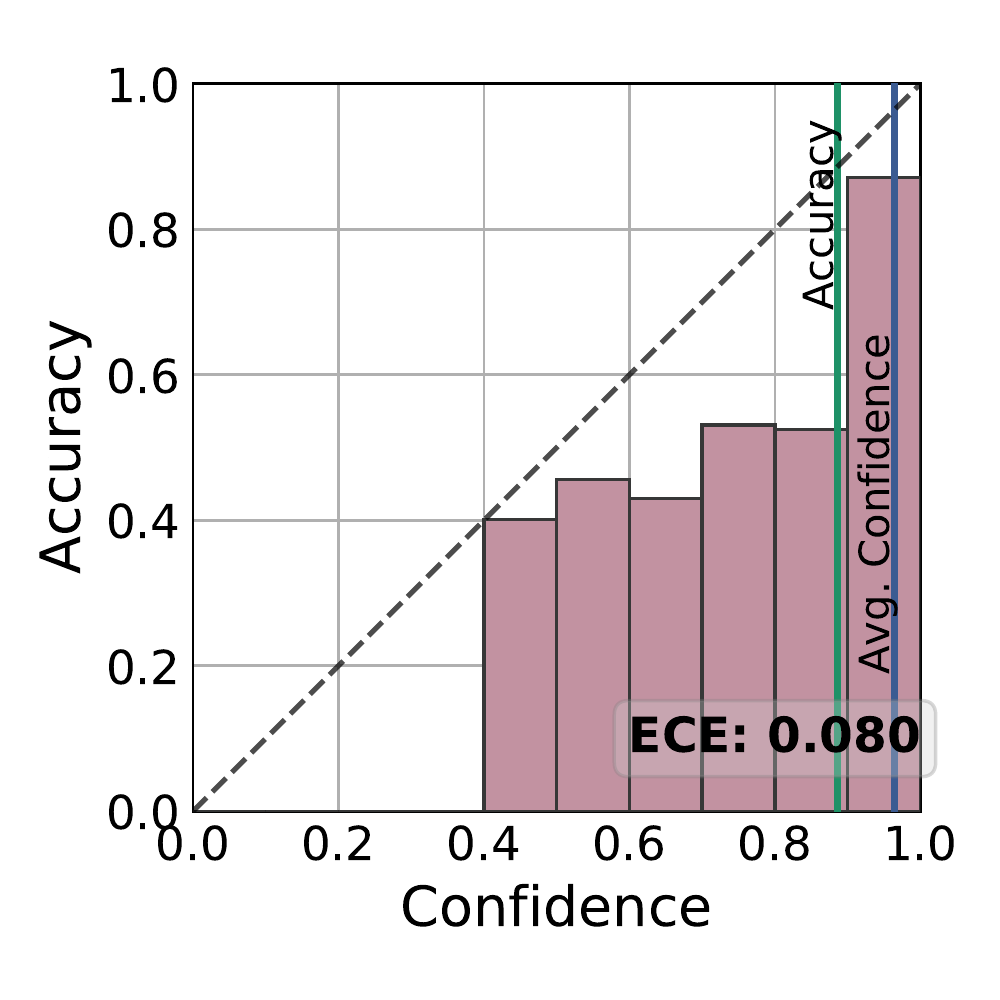}
    \vspace{-2em}
    \caption{192 steps}
    \end{subfigure}
    \caption{CIFAR10 image classification with a SDE-BNN. Generalization improves marginally compared to a trained model during inference in \ref{fig:stepstrained}, as tuning solver step size does not yield significant differences in calibration outcomes.}
    \label{fig:cifar10_calib_spectrum_inference}
\end{figure}
\vspace{-0.5cm
}
\subsection{SDE solver and adjoint settings}
\label{app:adjointadaptivesolver}

These were run with a SDE-BNN for MNIST image classification, to compare the performance and run-time cost across different solver settings. Comparably, backpropagation through the solver averaged 162.58 sec / epoch while the adjoint method averaged 135.90 sec / epoch in terms of wall clock time.
\begin{figure}[H] 
    \centering
    \begin{subfigure}[b]{0.32\linewidth}
    \includegraphics[width=\linewidth]{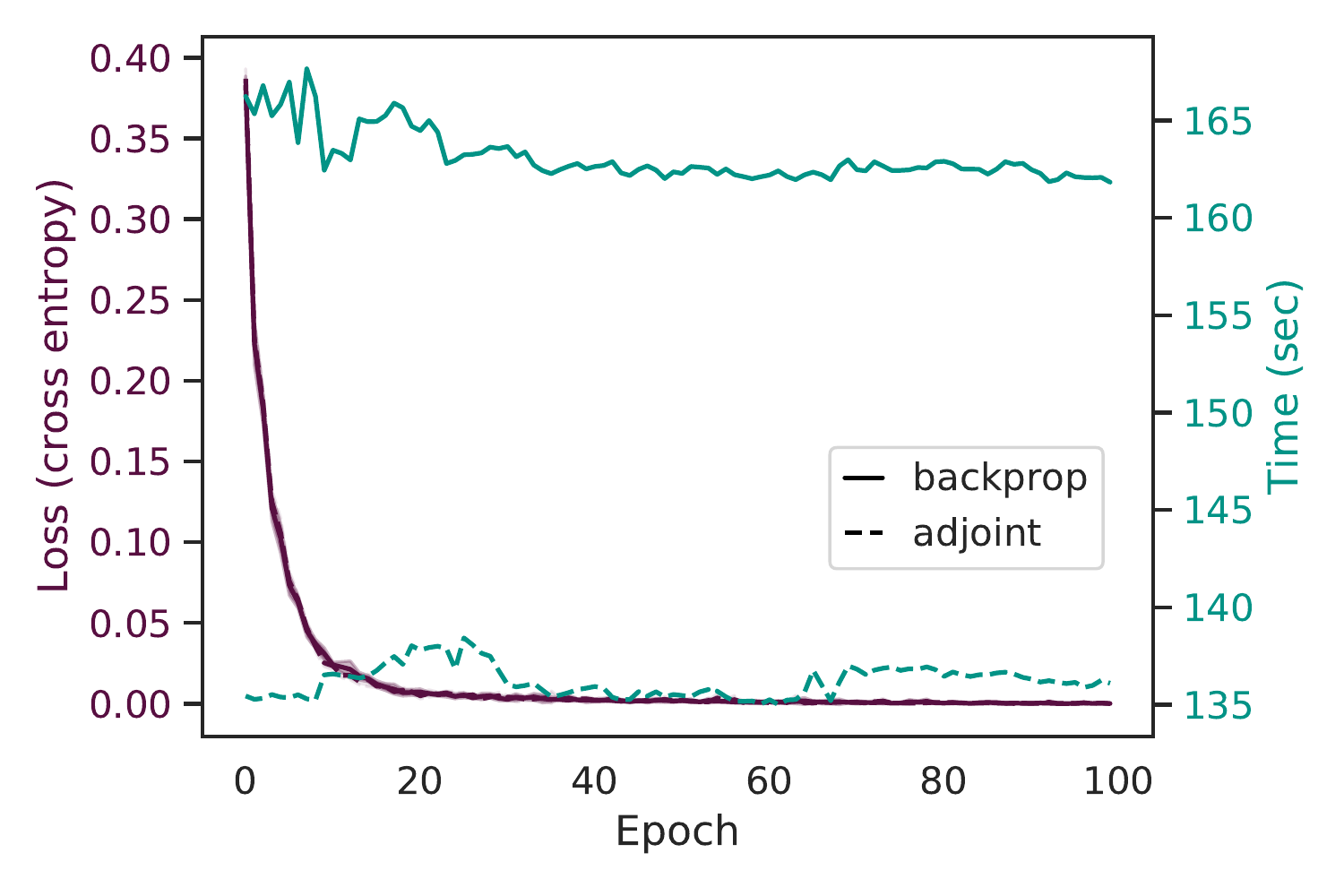}
    \vspace{-2em}
    \end{subfigure}
    \begin{subfigure}[b]{0.32\linewidth}
    \includegraphics[width=\linewidth]{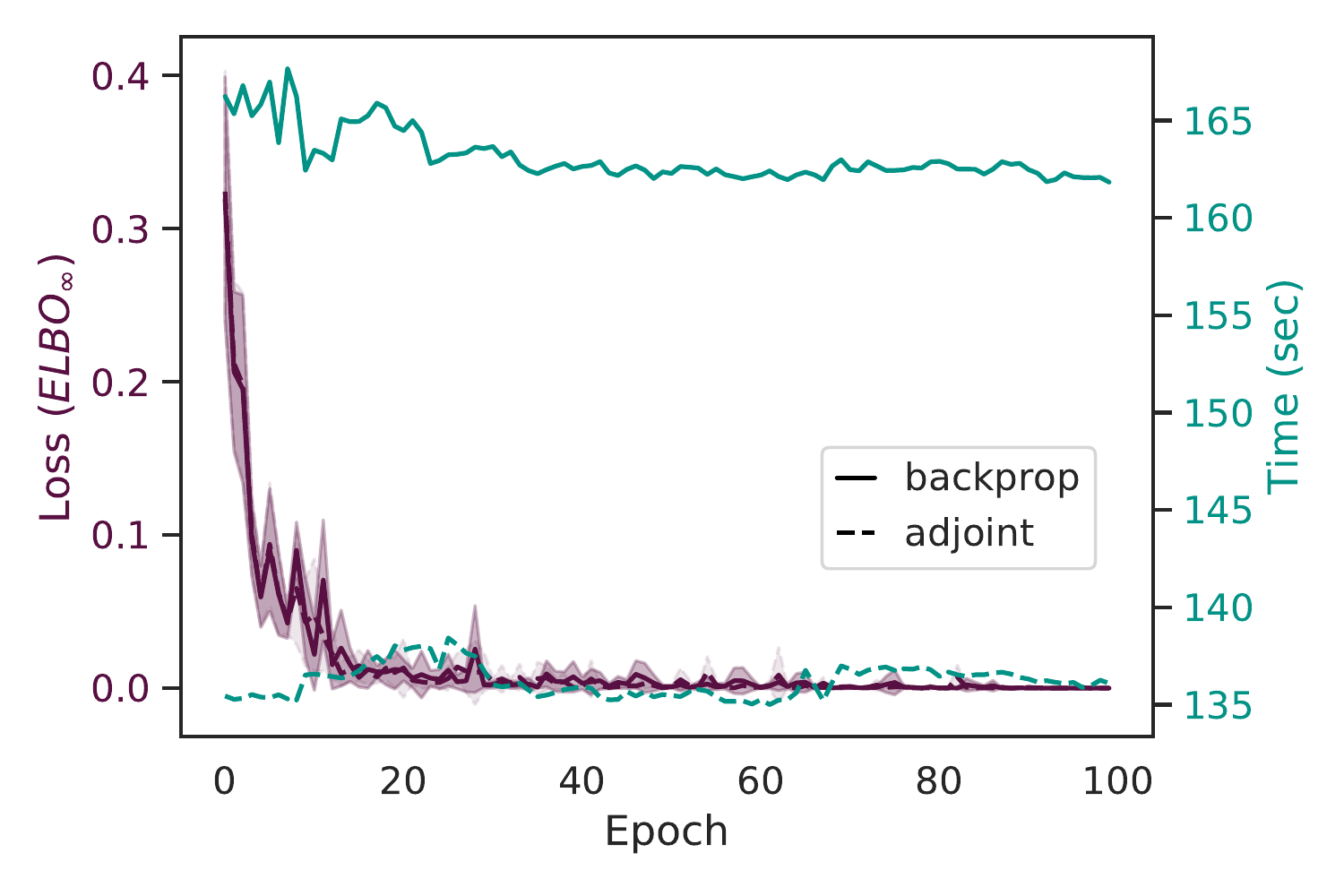}
    \vspace{-2em}
    \end{subfigure}
    \begin{subfigure}[b]{0.32\linewidth}
    \includegraphics[width=\linewidth]{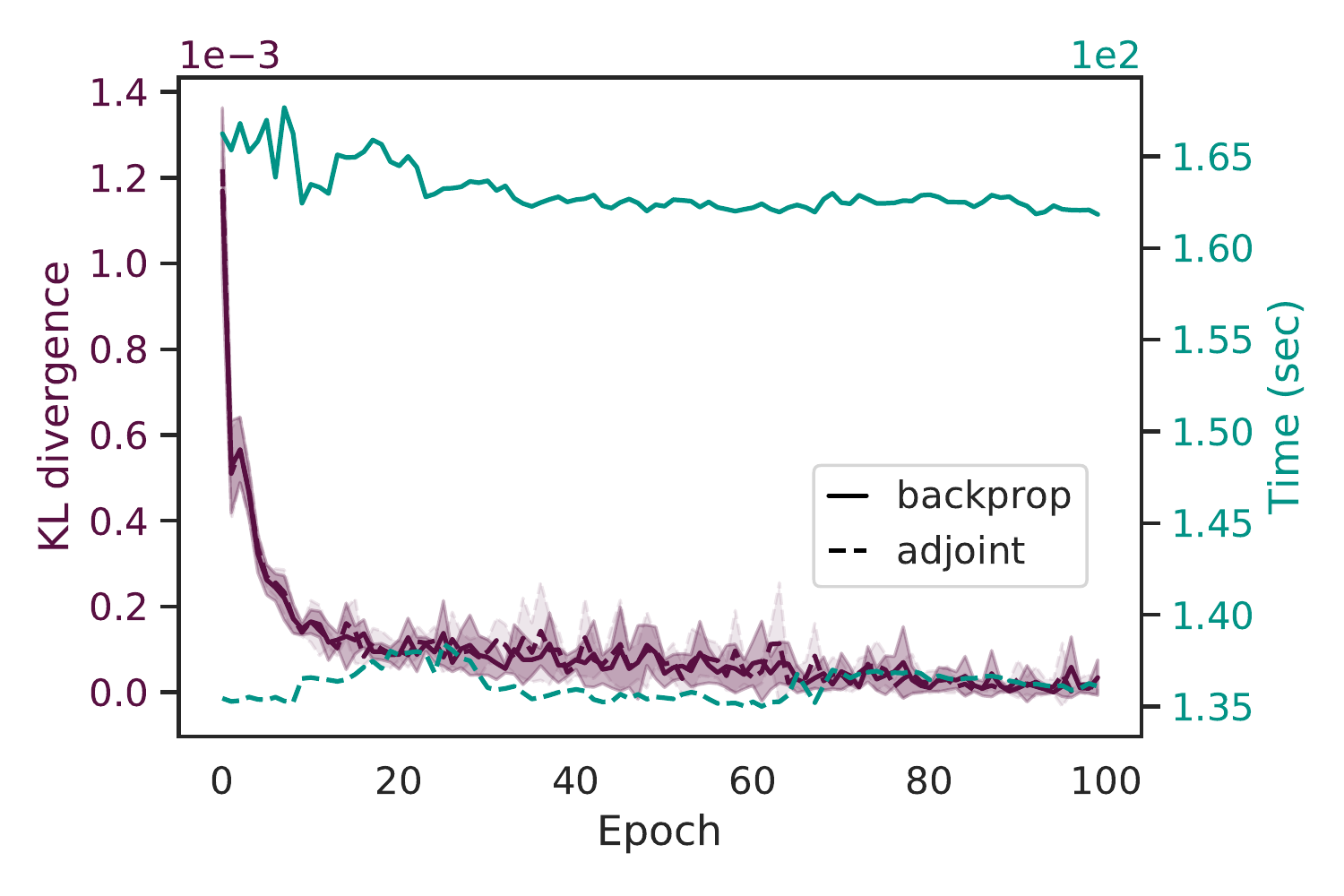}
    \vspace{-2em}
    \end{subfigure}
    \caption{Backpropagation through the SDE solver yields similar optimization dynamics but is less time efficient than the adjoint method.}
    \label{fig:mnist_adjoint}
\end{figure}
\vspace{-0.8cm}
\begin{figure}[H] 
    \centering
    \begin{subfigure}[b]{0.31\linewidth}
    \includegraphics[trim={0 0 2.15cm 0},clip,width=\linewidth]{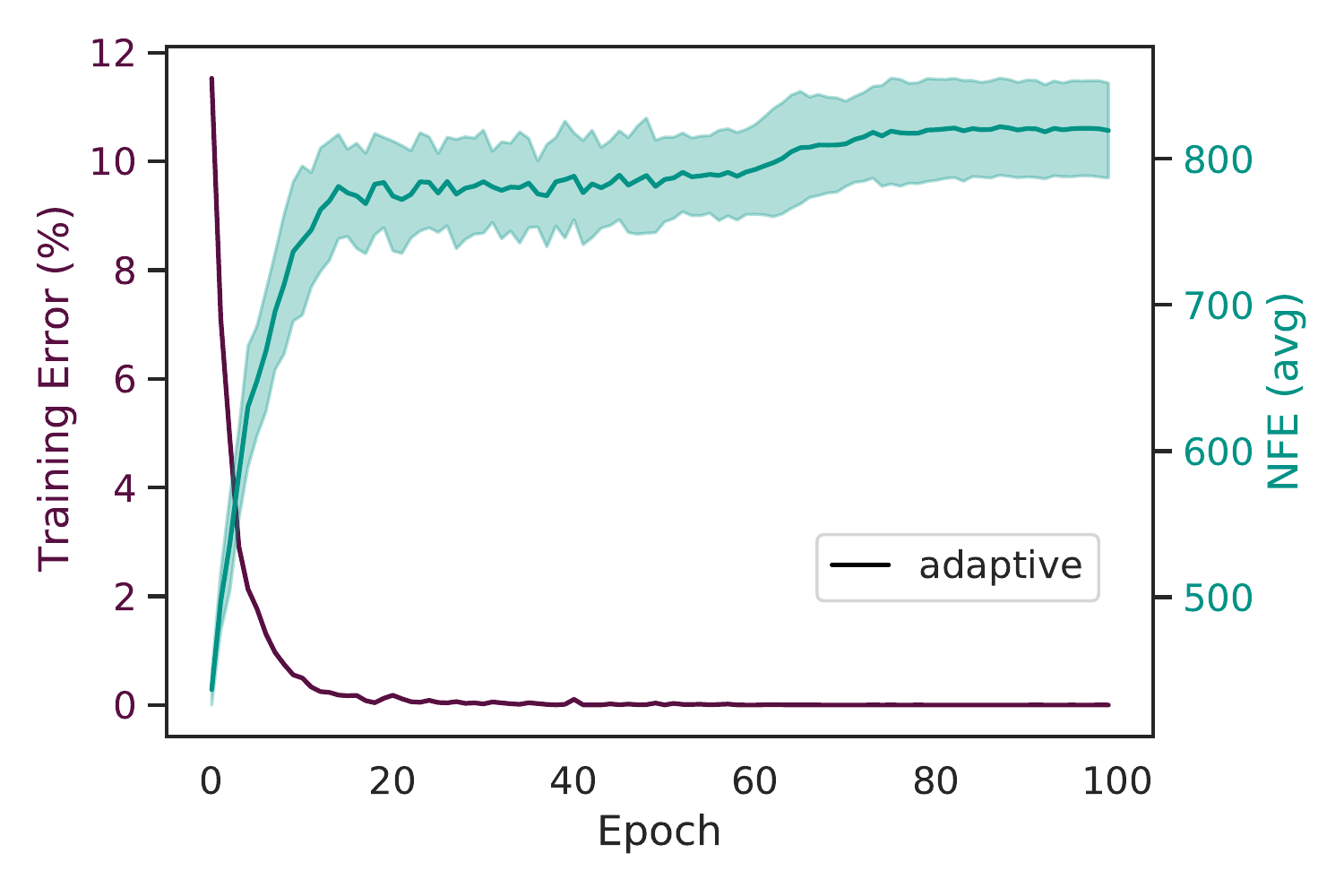}
    \vspace{-2em}
    \end{subfigure}
    \begin{subfigure}[b]{0.31\linewidth}
    \includegraphics[trim={0 0 2.15cm 0},clip,width=\linewidth]{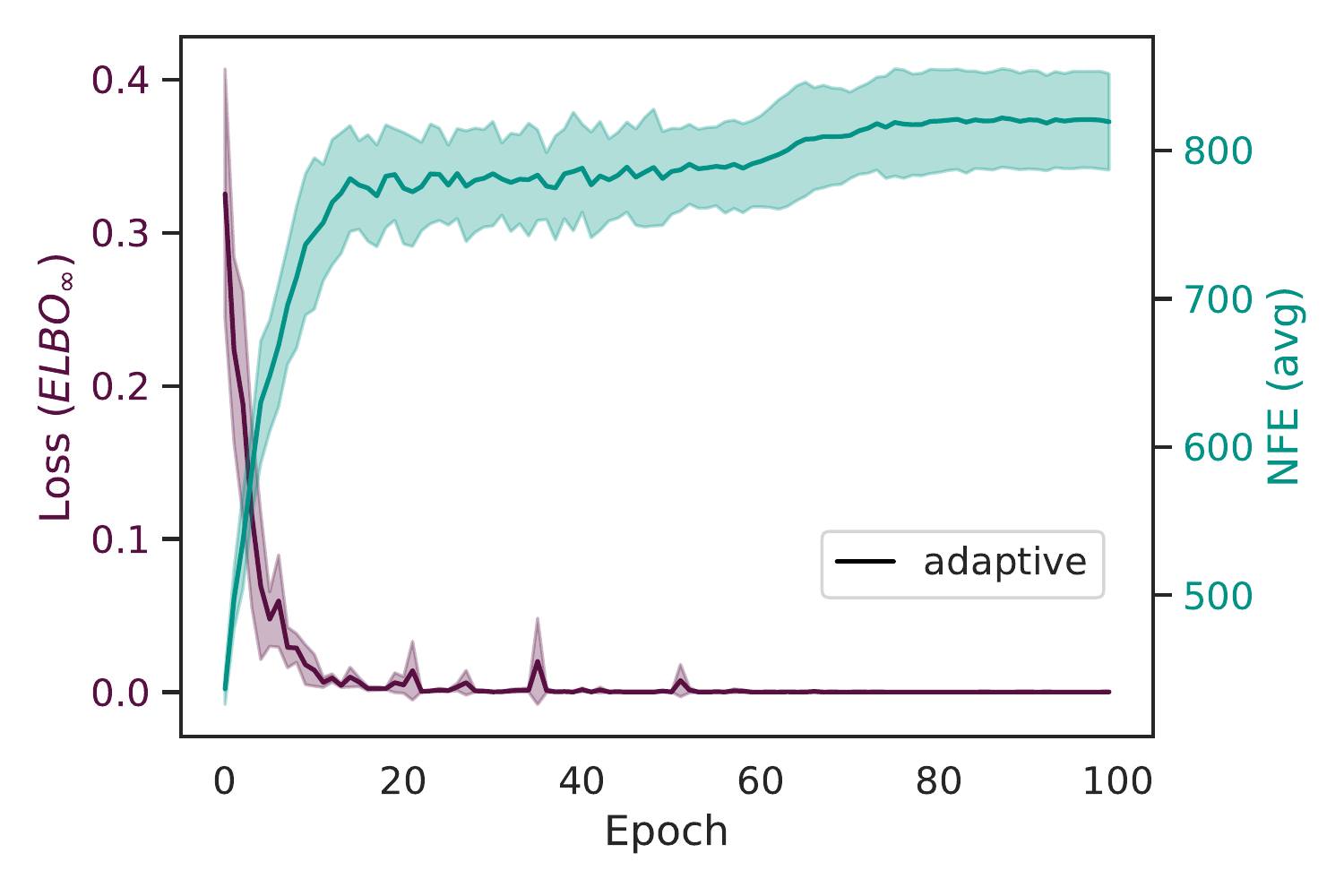}
    \vspace{-2em}
    \end{subfigure}
    \begin{subfigure}[b]{0.36\linewidth}
    \includegraphics[width=\linewidth]{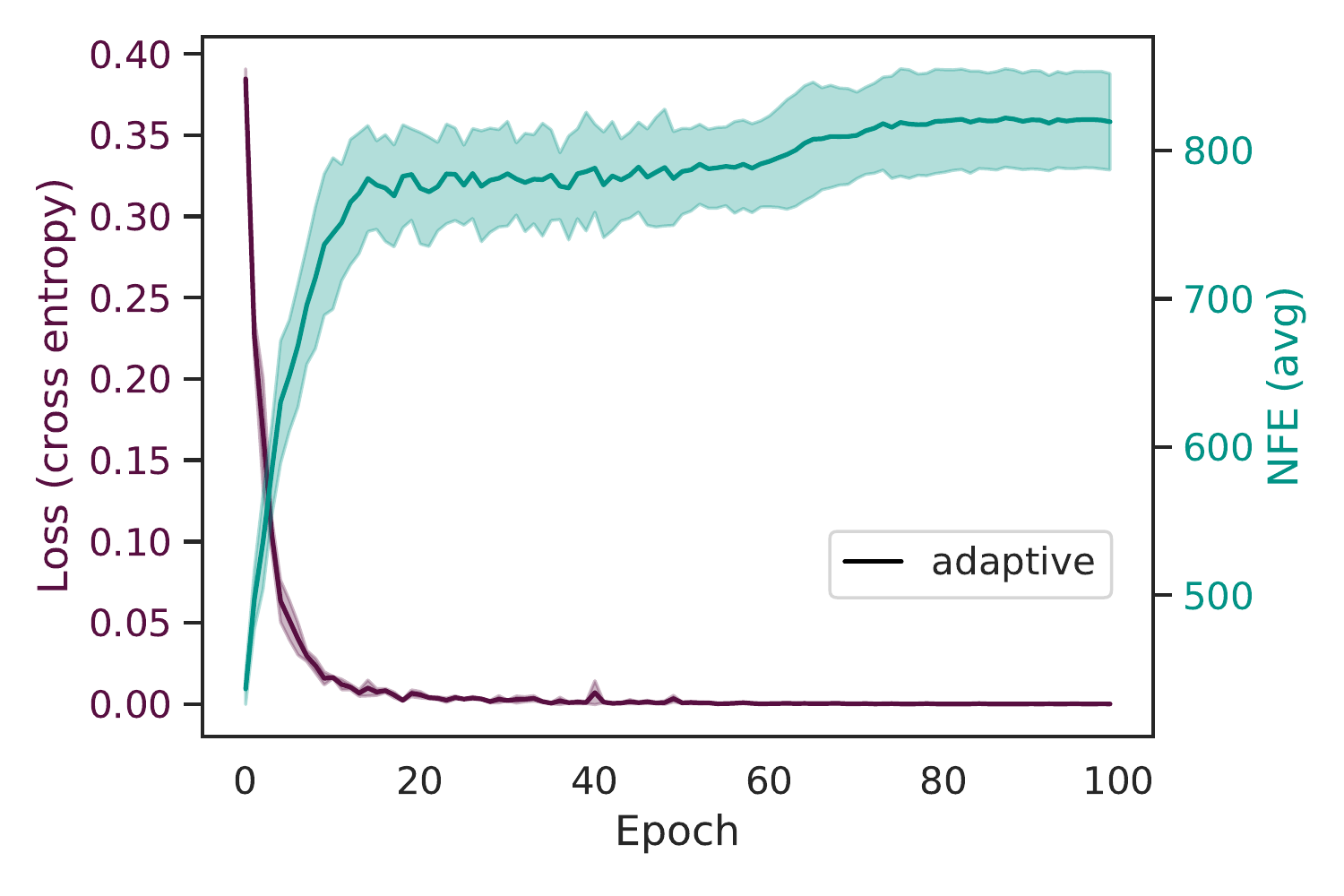}
    \vspace{-2em}
    \end{subfigure}
    \caption{Trade-off between solver speed and convergence during training. Adaptive refers to training with the stochastic adjoint in both forward and reverse modes here.}
    \label{fig:mnist_adaptive}
\end{figure}
\vspace{-0.75cm}
\begin{figure}[H]
    \centering
    \begin{subfigure}[b]{0.31\linewidth}
    \includegraphics[trim={0 0 2.15cm 0},clip,width=\linewidth]{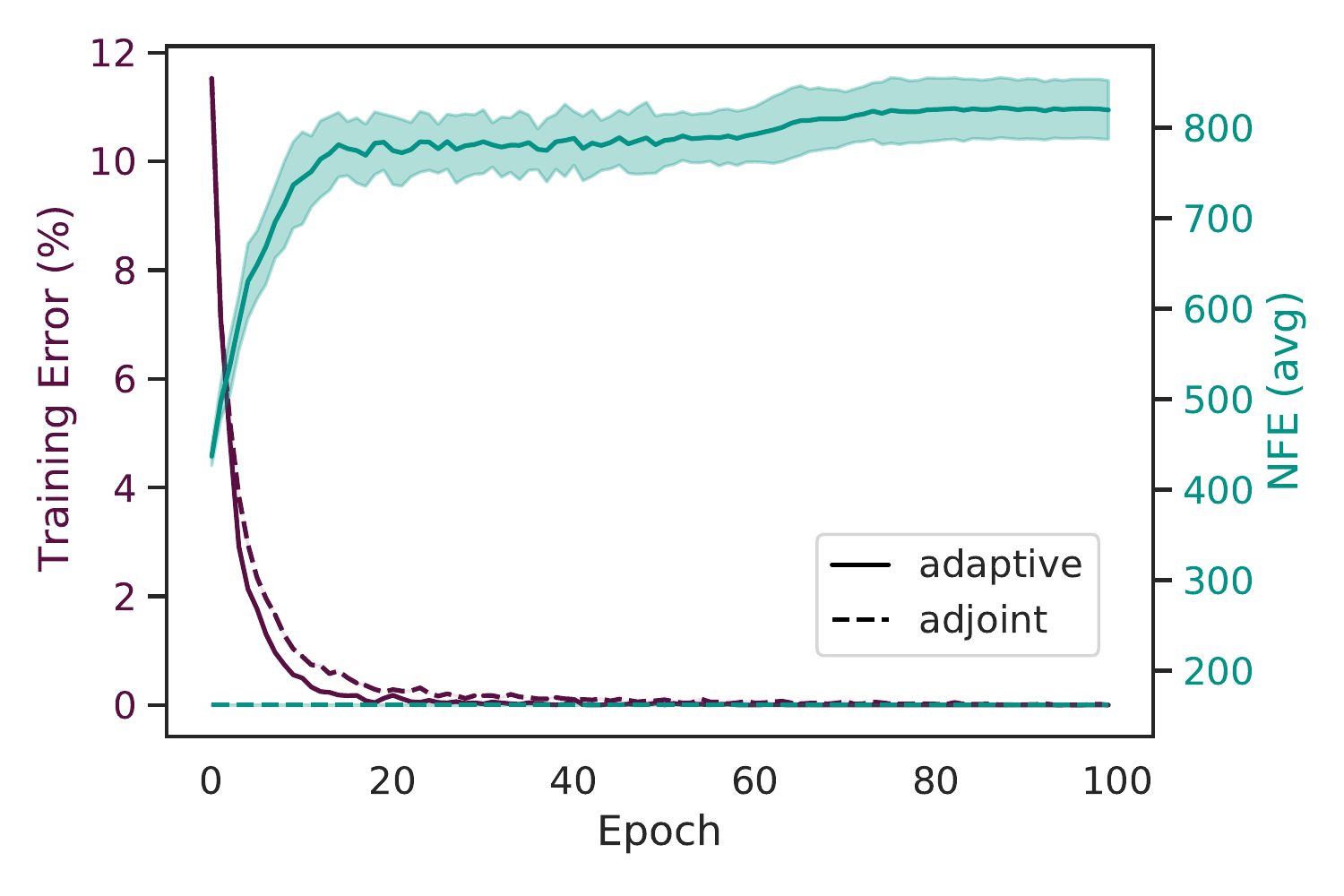}
    \vspace{-2em}
    \end{subfigure}
    \begin{subfigure}[b]{0.31\linewidth}
    \includegraphics[trim={0 0 2.15cm 0},clip,width=\linewidth]{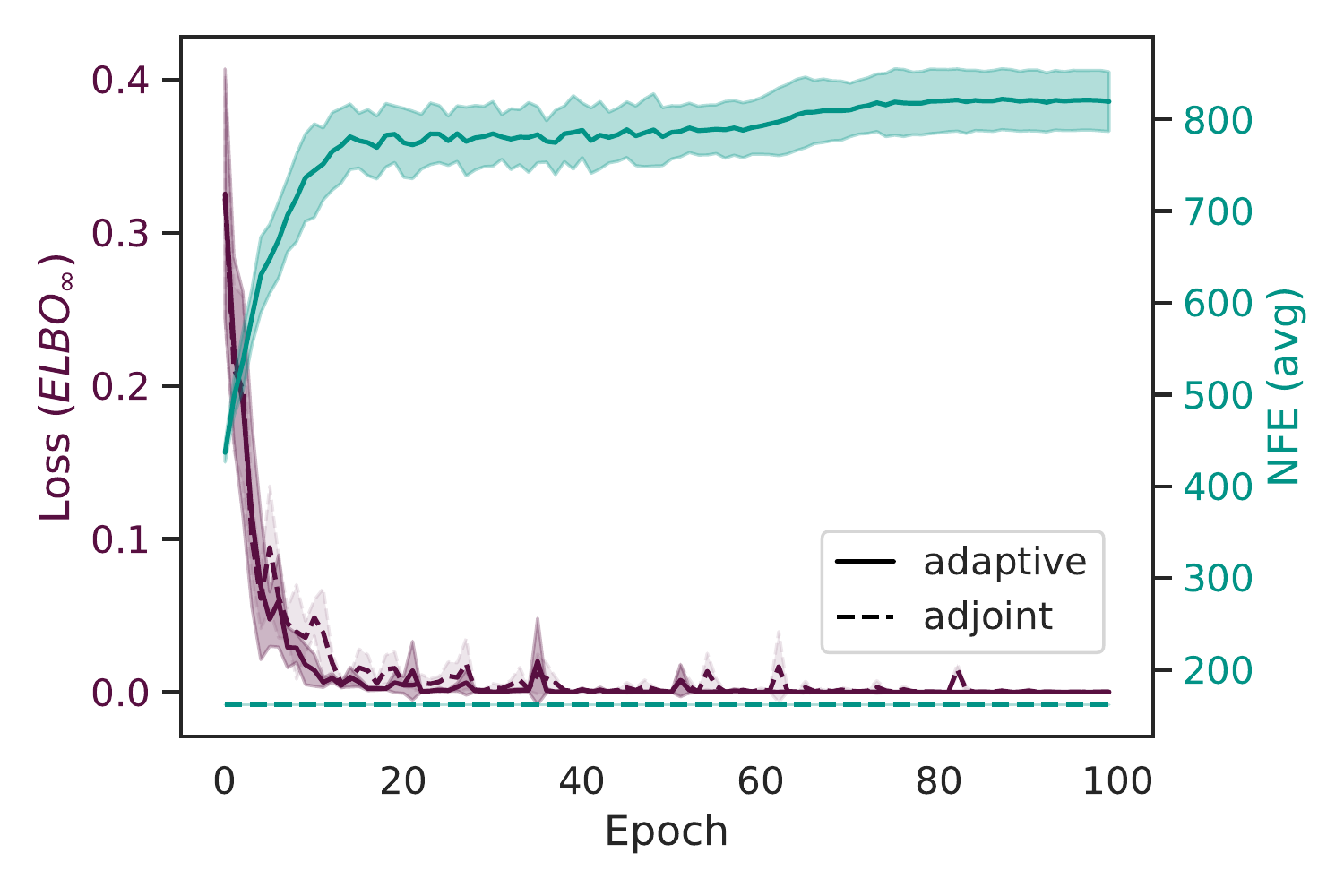}
    \vspace{-2em}
    \end{subfigure}
    \begin{subfigure}[b]{0.36\linewidth}
    \includegraphics[width=\linewidth]{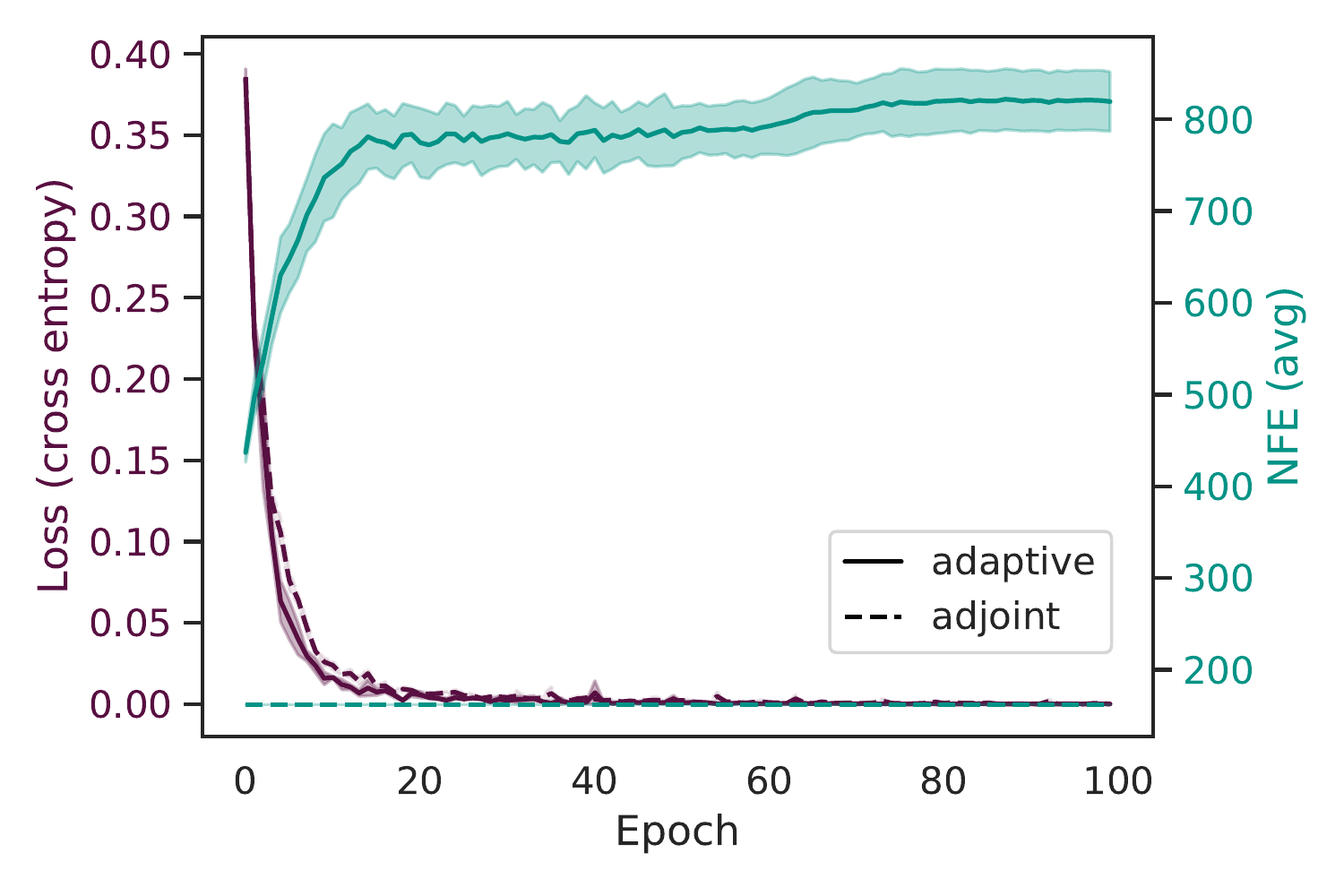}
    \vspace{-2em}
    \end{subfigure}
    \caption{Trade-off between solver speed and precision during training. Adaptive-order optimization trajectories were comparable to fixed-order solvers and were thus not applied to the classification tasks since computational resources were not under constraint.}
    \label{fig:mnist_adaptive_adjoint}
\end{figure}